\documentclass[10pt,twocolumn,letterpaper]{article}

\usepackage{iccv}
\usepackage{times}
\usepackage{epsfig}
\usepackage{graphicx}
\usepackage{amsmath}
\usepackage{amssymb}
\usepackage{url}
\usepackage{stackengine}
\usepackage{sidecap}
\usepackage{color}
\usepackage{subcaption}
\usepackage{enumitem}
\usepackage[normalem]{ulem}
\usepackage{multirow}
\usepackage{wrapfig}
\usepackage{pdflscape}
\usepackage{booktabs}
\usepackage{slashbox}
\usepackage{textcomp}
\usepackage[symbol]{footmisc}

\usepackage[pagebackref=true, breaklinks=true, letterpaper=true, colorlinks, bookmarks=false]{hyperref}

\iccvfinalcopy

\newcommand{\listenerA}{\textit{Baseline}}
\newcommand{\listenerB}{\textit{Early-Context}}
\newcommand{\listenerC}{\textit{Combined-Interpretation}}
\newcommand{\cic}{\textit{\textbf CiC}}

\usepackage{amsmath,amsfonts,bm}

\def\reals{{\mathbb R}}      % Real numbers

\def\eqref#1{equation~\ref{#1}}
\def\1{\bm{1}}

\DeclareMathAlphabet{\mathsfit}{\encodingdefault}{\sfdefault}{m}{sl}
\SetMathAlphabet{\mathsfit}{bold}{\encodingdefault}{\sfdefault}{bx}{n}

\def\delequal{\mathrel{\ensurestackMath{\stackon[1pt]{=}{\scriptscriptstyle\Delta}}}}

\DeclareSymbolFont{extraup}{U}{zavm}{m}{n}
\DeclareMathSymbol{\varheart}{\mathalpha}{extraup}{86}
\DeclareMathSymbol{\vardiamond}{\mathalpha}{extraup}{87}

\ificcvfinal\fi
\begin{document}

\title{ShapeGlot: Learning Language for Shape Differentiation}

\twocolumn[{
\renewcommand\twocolumn[1][]{#1}
\author{
  \vspace{-10pt}
  Panos Achlioptas\footnotemark
  \and
  Judy Fan
  \and
  Robert Hawkins
  \\
  \and
  \begin{tabular}[t]{c@{\extracolsep{2.0em}}c@{\extracolsep{2.0em}}c@{\extracolsep{2.0em}}c}    
    \multicolumn{3}{c}{Noah Goodman \qquad Leonidas Guibas}\\
    \multicolumn{3}{c}{Stanford University}    
  \end{tabular}       
}

\maketitle

 {
  \centering
  \vspace{-0.2cm}
  \centering
  \includegraphics[width=\textwidth]{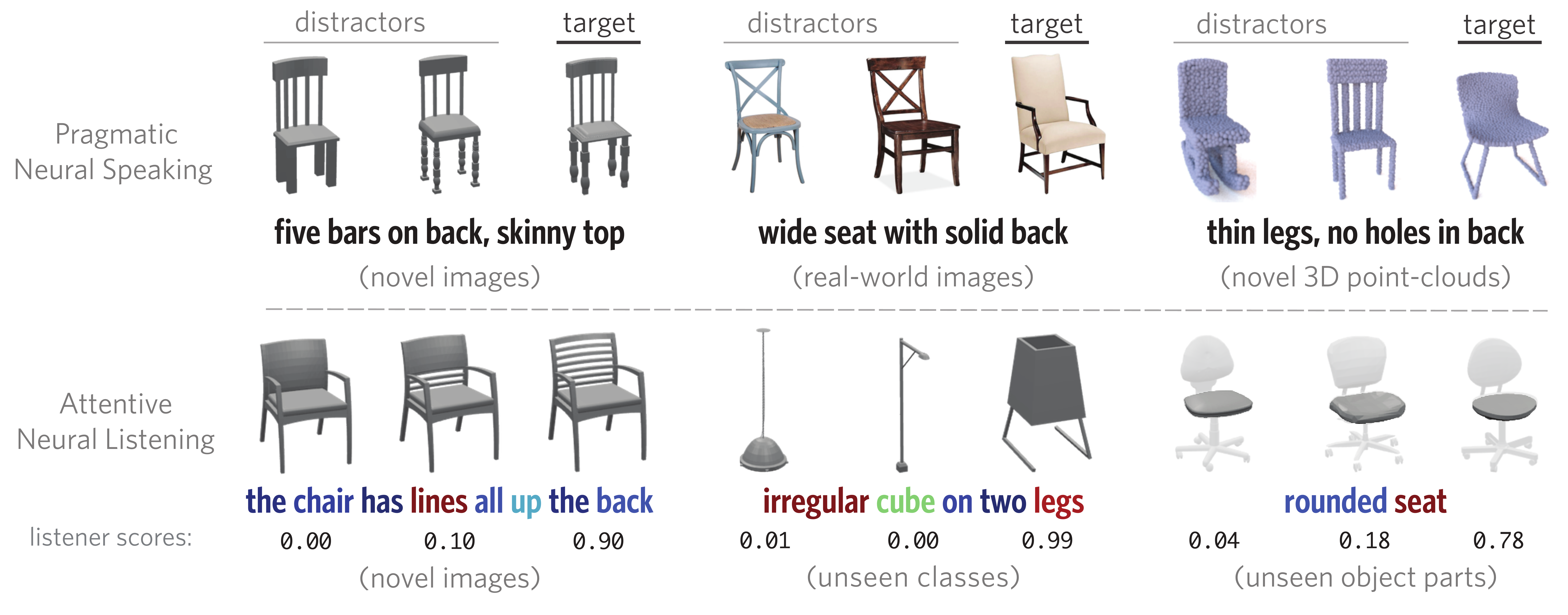}
  \captionof{figure}{
  We introduce a large corpus of utterances that refer to the shape of objects and develop neural speakers and listeners with broad generalization capacity. {\bf Top row}: A neural speaker generates utterances to distinguish a ``target" from two ``distractors" in \textit{unseen}: images of synthetic data (left), out-of-distribution (OOD) \textit{real-world} images (center), and 3D point-clouds of CAD models (right). {\bf Bottom row}: A neural listener interprets human-generated utterances for \textit{unseen} (left-to-right): images of synthetic data, OOD object \textit{categories} (here, lamps), and OOD object \textit{parts}. Listener scores indicate model interpretation about which object the utterance refers to. Words are color-coded according to the attention placed by the neural listener (warmer color indicates higher attention).}
  \label{fig:teaser}
 }
 \vspace{0.5cm}
}]

\footnotetext[1]{Corresponding author: \texttt{optas@cs.stanford.edu}\newline Project webpage: \url{https://www.bit.ly/shapeglot}}

%%%%%%%%% ABSTRACT
\begin{abstract}
 \qquad In this work we explore how fine-grained differences between the {\bf shapes} of common objects are expressed in language, grounded on {\bf images and 3D models} of the objects. We first build a large scale, carefully controlled dataset of human utterances that each refers to a 2D rendering of a 3D CAD model so as to distinguish it from a set of  shape-wise similar alternatives. Using this dataset, we develop neural language understanding (listening) and production (speaking) models that vary in their grounding (pure 3D forms via point-clouds vs. rendered 2D images), the degree of pragmatic reasoning captured (e.g. speakers that reason about a listener or not), and the neural architecture (e.g. with or without attention). We find models that perform well with both synthetic and human partners, and with held out utterances and objects. We also find that these models are amenable to {\bf zero-shot} transfer learning to novel object classes (e.g. transfer from training on chairs to testing on lamps), as well as to real-world images drawn from furniture catalogs. Lesion studies indicate that the neural listeners depend heavily on part-related words and associate these words correctly with {\bf visual parts} of objects (without any explicit network training on object parts), and that transfer to novel classes is most successful when known part-words are available. This work illustrates a practical approach to language grounding, and provides a case study in the relationship between object shape and linguistic structure when it comes to {\bf object differentiation}.
\end{abstract}

\section{Introduction}
Objects are best understood in terms of their structure and function, both of which are built on a foundation of object parts and their relations \cite{Pictorial_Structures, deformable_part_models, Latent_objects, dubrinova_parts}. Natural languages have been optimized across human history to solve the problem of efficiently communicating the aspects of the world most relevant to one's current goals \cite{kirby2015compression,GibsonEtAl17_ColorNamingUse}. As such, languages can provide an effective medium to describe the \textit{shapes} and the \textit{parts} of different objects, and to express \textit{object differences}.  For example, when we see a chair we can decompose it into semantically meaningful parts, like a \textsl{back} and a \textsl{seat}, and can combine words to create utterances that reflect their geometric and topological \textit{shape-properties} e.g.~`wide seat with a solid back’. Moreover, given a specific communication context, we can craft references that are not merely true, but which are also relevant: i.e.~we can refer to the lines found in a chair's back to \textit{distinguish} it among other similar objects (see Fig.~\ref{fig:teaser}).

In this paper we explore this interplay between natural, referential language, and the shape of common objects. While a great deal of recent work has explored visually-grounded language understanding \cite{referit, nagaraja16, licheng_16, luo2017comprehension, baby_talk, mattnet}, the resulting models have limited capacity to reflect the geometry and topology (i.e. the shape) of the underlying objects. This is because reference in previous studies was possible using properties like \textit{color}, or properties regarding the object and it's hosting environment (e.g.~it's absolute, or relative to other objects, \textit{location}). Indeed, eliciting natural language that refers only to shape properties requires carefully controlling the objects, their presentation, and the linguistic task. To address such challenges, we use pure 3D representations of objects (CAD models), which allow for flexible and \textit{controlled} presentation (i.e.~textureless, uniform-color objects, viewed without obstruction in a fixed pose). We further make use of the 3D form to construct a reference game task in which the referred object is similar \textit{shape-wise} to the contrasting objects.
The result of this effort is a new multimodal dataset, termed \textit{{\textbf CiC}} (\textit{{\textbf C}hairs {\textbf i}n {\textbf C}ontext}), comprised of 4,511 unique chairs from ShapeNet~\cite{shapenet} and 78,789 referential utterances. In CiC chairs are organized into 4,054 sets of size 3 (representing contrastive communication contexts) and each utterance is intended to distinguish a chair in context. The visual differences among the grouped objects require a deep understanding of very fine-grained shape properties (especially, for \textit{Hard} contexts, see Section~\ref{sec:dataset}); the language that people use to do so is correspondingly complex, exhibiting rich compositionality.

We use CiC to train and analyze a variety of modern neural language understanding (listening) and production (speaking) models. These models vary in their grounding (pure 3D forms via point-clouds vs. rendered 2D images), the degree of pragmatic reasoning captured (e.g. speakers that reason about a listener or not) and the neural architecture (e.g. with or without word attention, and with context-free or context-aware object encoders). We evaluate these models on the original reference game task with both synthetic and human partners, and with held out utterances and objects, finding strong performance. Since language conveys abstractions, such as object parts, that are shared between object categories, we hypothesized that our models learn \textit{robust} representations that are transferable to objects of unseen classes (e.g.~training on chairs while testing on lamps). Indeed, we show that these models have strong generalization capacity to novel object \textit{categories}, as well as to \textit{real-world} colored images drawn from furniture catalogs.

Finally, we explore \textit{how} our models are succeeding on these communication tasks. We demonstrate that the neural listener learns to prioritize the same abstractions in objects (i.e.~properties of chair parts) that humans do in solving the communication task, despite \textit{never} being provided with an explicit decomposition of these objects into parts. Similarly, we show that transfer learning to novel object classes is most successful when known part-related words are available. Last, we show that a neural speaker that is \emph{pragmatic}---planing utterances in order to convey the right target object to an imagined listener
---produces significantly more informative utterances than a \emph{literal} (listener-unaware) speaker, as measured by human performance in identifying the correct object.

\section{Dataset and task}
\label{sec:dataset}
\begin{wrapfigure}{r}{0.40\columnwidth}
\centering
\includegraphics[width=0.40\columnwidth]{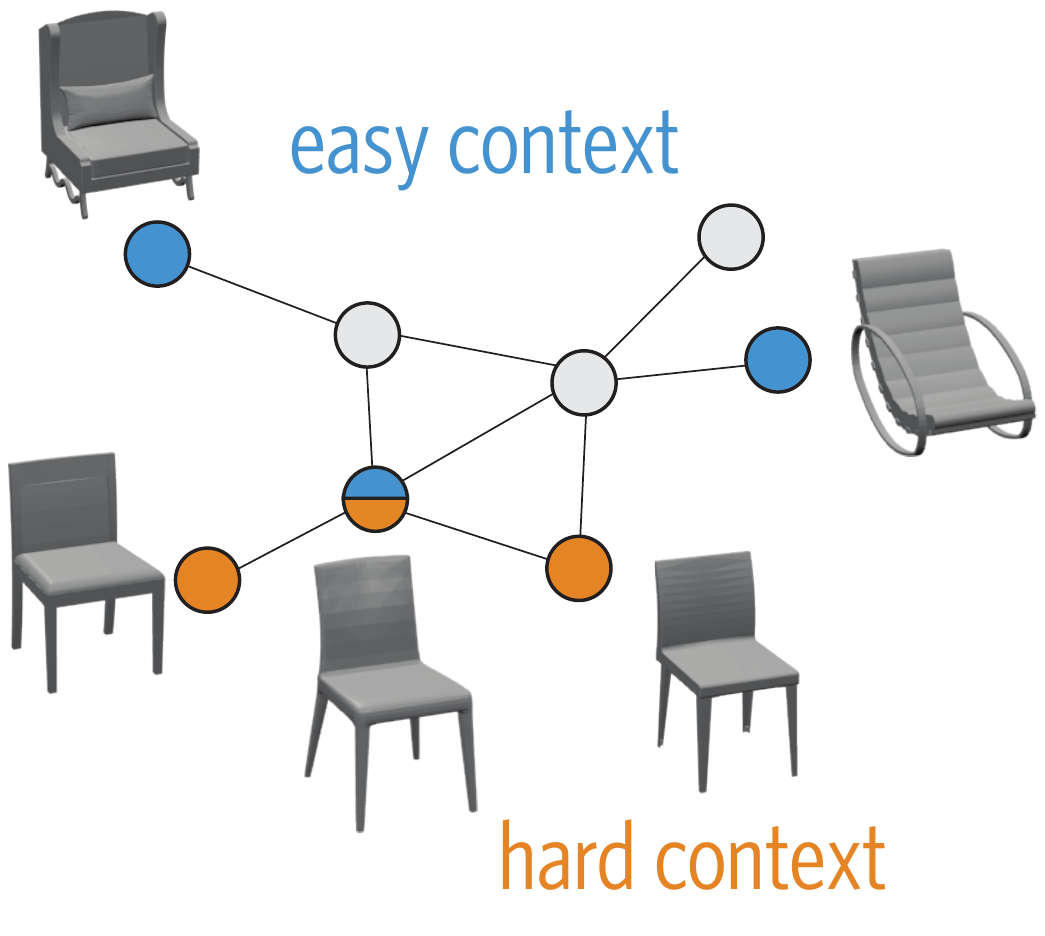}
\vspace{-14pt}
\label{fig:contexts}
\end{wrapfigure}
\textit{{\textbf CiC}} (\textit{{\textbf C}hairs {\textbf i}n {\textbf C}ontext}) consists of triplets of chairs coupled with referential utterances that aim to distinguish one chair (the ``target") from the remaining two (the ``distractors"). To obtain such utterances, we paired participants from Amazon's Mechanical Turk (AMT) to play an online reference game \cite{Hawkins15_RealTimeWebExperiments}. On each round of the game, the two players were shown the same triplet of chairs. The designated target chair was privately highlighted for one player (the ``speaker'') who was asked to send a message through a chat box such that their partner (the ``listener'') could successfully select it from the context. To ensure speakers used {\em only} shape-related information, we scrambled the positions of the chairs for each participant independently and used textureless, uniform-color renderings of pre-aligned 3D CAD models, taken from the same viewpoint. To ensure communicative interaction was natural, no constraints were placed on the chat box: referring expressions from the speaker were occasionally followed by clarification questions from the listener or other discourse. 

A key decision in building our dataset concerned the construction of contexts that would reliably elicit {\em diverse} and potentially {\em  very} fine-grained contrastive language. To achieve diversity we considered all ${\sim}$7,000 chairs from ShapeNet. This object class is geometrically complex, highly diverse, and abundant in the real world. To control the granularity of fine-grained distinctions that were necessary in solving the communication task, we constructed two types of contexts: \textit{Hard} contexts consisted of very similar shape-wise chairs, and \textit{Easy} contexts consisted of less similar chairs. To measure shape-similarity in an unsupervised manner, we used the latent space derived from an Point Cloud-AutoEncoder (PC-AE) \cite{achlioptas2018latent_pc}. We note, that point-clouds are an intrinsic representation of a 3D object, \textit{oblique} to color or texture. After extracting a 3D point-cloud from the surface of each ShapeNet model we computed the underlying K-nearest-neighbor graph among all models according to their PC-AE embedding distances. 
For a chair with sufficiently high-in degree on this graph (corresponding intuitively to a canonical chair) we contrasted it with four distractors: the two \textit{closest} to it in latent-space, and two that were sufficiently far (see inset for a demonstration and at the Appendix for additional details). Last, to reduce potential data biases we \textit{counterbalanced} each communication context, by considering every chair of a given context as target, in at least four games.

Before presenting our neural agents, we identify some distinctive properties of our corpus.  Human performance on the reference game was high, but listeners made significantly more errors in the Hard triplets (accuracy $94.2\%$ vs. $97.2\%, z = 13.54, p < 0.001$). Also, in Hard triplets longer utterances were used to describe the targets (on average 8.4 words vs. 6.1, $t = -35, p <0.001$). A wide spectrum of descriptions was elicited, ranging from the more holistic/categorical (e.g.~ ``the rocking chair'') common for Easy triplets, to more complex and fine-grained language, (e.g.~ ``thinner legs but without armrests'') common for Hard triplets. Interestingly, 78\% of the utterances used at least one part-related word: ``back'', ``legs'', ``seat,'' ``arms'', or closely related synonyms e.g.~ ``armrests''.
\section{Neural listeners}
\label{sec:neural_listeners}
Constructing neural listeners that reason effectively about shape properties is a key contribution of our work. % It lays the foundation for creating speakers that utter discriminative utterances and enables the creation of an object retrieval system that operates with linguistic queries.
Below we conduct a detailed comparison between three distinct architectures, highlight the effect of different regularization techniques, and investigate the merits of different representations of 3D objects for the listening task, namely, 2D rendered images and 3D surface point clouds. In what follows, we denote
the three objects of a communication context as $O=\{o_1, o_2, o_3\}$, the corresponding word-tokenized utterance as $U = u_1, u_2,\ldots$ and as $t \in O$ the designated target.

Our proposed listener is inspired by \cite{monroe_colors}. It takes as input a (latent code) vector that captures shape information for each of the objects in $O$, and a (latent code) vector for each token of $U$, and outputs an object--utterance compatibility score $\mathcal{L}(o_{i}, U) \in [0, 1]$ for each input object. At its core lies a multi-modal LSTM~\cite{hochreiter1997long} that receives as input (``is grounded" with) the vector of a single object, processes the word-sequence $U$, and is read out by a final MLP to yield a single number (the compatibility score). This is repeated for each $o_{i}$, {\em sharing} all network parameters across the objects. The resulting three scores are soft-max normalized and compared to the ground-truth indicator vector of the target under the cross-entropy loss.\footnote{Architecture details, hyper-parameter search strategy, and optimal hyper-parameters for all experiments are described in the Appendix.}

\textbf{Object encoders} \hspace{3mm}
We experimented with three object representations to capture the underlying shapes: (a) the bottleneck vector of a pretrained Point Cloud-AutoEncoder (PC-AE), (b) the embedding provided by a convolutional network operating on single-view images of non-textured 3D objects, or (c) a combination of (a) and (b). Specifically, for (a) we use the PC-AE architecture of~\cite{achlioptas2018latent_pc} trained with single-class point clouds extracted from the surfaces of 3D CAD models, while for (b) we use the activations of the penultimate layer of a VGG-16~\cite{simonyan2014very}, pre-trained on ImageNet~\cite{deng2009imagenet}, and fine-tuned on an 8-way classification task with images of objects from ShapeNet. For each representation we project the corresponding latent code vector to the input space of the LSTM using a fully connected (FC) layer with $L_2$-norm weight regularization. The addition of these projection-like layers improves the training and convergence of our system.

While there are many ways to simultaneously incorporate the two modalities in the LSTM, we found that the best performance resulted when we ground the LSTM with the image code, concatenate the LSTM's final output (after processing $U$) with the point cloud code, and feed the concatenated result in a shallow MLP to produce the compatibility score. We note that grounding the LSTM with point clouds and using images towards the end of the pipeline, resulted in a significant performance drop (${\sim}4.8\%$ on average). Also, proper regularization was \textit{critical}: adding dropout at the input layer of the LSTM and $L_2$ weight regularization and dropout at and before the FC projecting layers improved performance ${\sim}10\%$. The token codes of each sentence where initialized with the GloVe embedding \cite{pennington2014glove} and fine-tuned for the listening task.

% We note that the addition of the MLP is slightly preferable to directly feeding the point-based codes in the LSTM ($\sim1\%$). However, the order of feeding the input is important. Grounding with point clouds and using images towards the end resulted in a significant deterioration of the performance ($\sim 6.5\%$). Last, using each modality in lieu of each other (e.g. feeding it twice) doesn't improve over a system that uses a single modality - implying a tangible (and as shown later, exploitable) complementarity between the two encodings.
% We hypothesize that the coarser point-based signal is not sufficient for fully deciphering (and thus grounding) the detailed geometric information conveyed by the language and as such is better used as an 'secondary' signal, and closer to the loss. 

\textbf{Incorporating context information} \hspace{3mm}
Our proposed baseline listener architecture (\listenerA , just described) first scores each object \textit{separately} then applies softmax normalization to yield a score distribution over the three objects. 
We also consider two alternative architectures that explicitly encode information about the {\em entire} context while scoring an object.
The first alternative (\listenerB), is identical to the proposed architecture, except for the codes used to ground the LSTM. Specifically, if $v_i$ is the image code vector of the i-th object ($o_i \in O$) resulting from VGG, instead of using $v_i$ as the grounding vector of $o_i$, a shallow convolutional network is introduced. This network, of which the output {\em is} the grounding code, receives the signal $f(v_j, v_{k}) || g(v_j, v_{k}) || v_i$, where $f, g$ are the symmetric (and norm-normalized), max-pool and mean-pool functions, $||$ denotes feature-wise concatenation and $v_j, v_{k}$ the alternative constrastive objects. Here, we use symmetric functions to induce that object-order is irrelevant for our task.
The second alternative architecture (\listenerC) first feeds the image vectors for \emph{all} three objects sequentially to the LSTM as inputs and then proceeds to process the tokens of $U$ {\em once}, to yield the three scores. Similarly to the \listenerA \ architecture, point clouds are incorporated in both alternatives via a separate MLP after the LSTM.

\textbf{Word attention} \hspace{3mm}
We hypothesized that a listener forced to prioritize a few words in each utterance would learn to prioritize words that express properties that distinguish the target from the distractors (and, thus, perform better). To test this hypothesis, we augment the listener models with a standard {\em bilinear attention mechanism} \cite{attention_rnn_clf}. Specifically, to estimate the ``importance" of each text-token $u_i$ we compare the output of the LSTM at $u_i$ (denoted as $r_i$) with the hidden state after the entire utterance has been processed (denoted as $h$). The relative importance of each word is $a_{i} \delequal r_i^{T} \times W_{\text{att}} \times h$, where $W_{\text{att}}$ is a trainable diagonal matrix. The final output of the LSTM uses this attention to combine all latent states: $\sum_{i=1}^{|U|} r_{i} \odot \hat{a}_{i}$, where $\hat{a}_{i} = \frac{\exp \left(a_i\right)}{\sum_j^{|U|} \exp\left( a_j\right)}$ and $\odot$ is the point-wise product. 

% The optimal parameters of each listener (and speaker), the hyper-parameter search strategy, and the exact details of training are provided in the Appendix~\ref{apndx:listener_details} and \ref{apndx:speaker_details}.
\section{Listener experiments}
\label{sec:listening_results}

%!TEX root = ../sections/main.tex
\begin{table*}[tbh!]
\centering
\begin{tabular}{c| c | c c c}
\hline
\multirow{2}{*}{\textbf{Architecture}} & 
	& \multicolumn{3}{c}{\textbf{Subpopulations}}  \\
	& \textbf{Overall} & \textbf{Hard} & \textbf{Easy} & \textbf{Sup-Comp} \\
\hline

\listenerC
&$75.9 \pm 0.5\%$
&$67.4 \pm 1.0\%$
&$83.8 \pm 0.6\%$
&$74.4 \pm 1.5\%$\\

\listenerB
&$79.4 \pm 0.8\%$
&$\textbf{70.1} \pm 1.3\%$
&$88.1 \pm 0.6\%$
&$75.6 \pm 2.2\%$\\

\listenerA
&$\textbf{79.6} \pm 0.8\%$
&$69.9 \pm 1.3\%$
&$\textbf{88.8} \pm 0.4\%$
&$\textbf{76.3} \pm 1.3\%$\\
\hline
\end{tabular}
\caption{Comparing different ways to include context. The simplest \listenerA \ model performs as well as more complex alternatives. Subpopulations are the subset of test data containing: hard trials (shape-wise similar distractors), easy trials, superlatives or comparatives.}
\label{table:listener_context_ablations}
\end{table*}
%!TEX root = ../sections/main.tex

\begin{table}[t!]
\centering
\begin{tabular}{c c c c}
\hline
& \begin{tabular}{@{}c@{}}\textbf{Input}    \\ \textbf{Modality}\end{tabular} & 
  \begin{tabular}{@{}c@{}}\textbf{Language} \\ \textbf{Task}    \end{tabular} & 
  \begin{tabular}{@{}c@{}}\textbf{Object}   \\ \textbf{Task}    \end{tabular} 
\\ \hline

\multirow{3}{*}{\textbf{\begin{tabular}[c]{@{}l@{}}No\\ Attention\end{tabular}}}

& Point Cloud	&$67.6 \pm 0.3\%$ & $66.4 \pm 0.7\%$ \\
& Image   		&$81.2 \pm 0.5\%$ & $77.4 \pm 0.7\%$ \\
& Both  		&$83.1 \pm 0.4\%$ & $78.9 \pm 1.0\%$ \\
\hline

\multirow{3}{*}{\textbf{\begin{tabular}[c]{@{}l@{}}With\\ Attention\end{tabular}}}
& Point Cloud   & $67.4 \pm 0.3\%$ & $65.6 \pm 1.4\%$ \\
& Image         & $81.7 \pm 0.5\%$ & $77.6 \pm 0.8\%$ \\
& Both  		& $\textbf{83.7} \pm 0.3\%$ & $\textbf{79.6} \pm 0.8\%$ \\
\hline
\end{tabular}
\caption{Performance of the \listenerA \ listener architecture using different object representations and with/without word level attention, in two generalization tasks.}
\label{main_listener_performance}
\end{table}

We begin our evaluation of the proposed listeners using two reference tasks based on different data splits. 
In the \emph{language generalization} task, we test on target objects that were seen as targets in at least one context during training but ensure that all utterances in the test split are from unseen speakers. 
In the more challenging \emph{object generalization} task, we restrict the set of objects that appeared as targets in the test set to be \emph{disjoint} from those in training such that all speakers \emph{and} objects in the test split are new.
For each of these tasks, we evaluate choices of input modality and word attention, using $[80\%,10\%,10\%]$ of the data, for training, validating and testing purposes. 

\listenerA \ listener accuracies are shown in Table \ref{main_listener_performance}.\footnote{In all results mean accuracies and standard errors across 5 random seeds are reported, to control for the data-split populations and the initialization of the neural-network.}
Overall the model achieves good performance. As expected, all listeners have higher accuracy on the language generalization task ($3.2\%$ on average). 
The attention mechanism on words yields a mild performance boost, as long as images are part of the input. 
Interestingly, images provide a significantly better input than point-clouds when only one modality is used. 
This may be due to the higher-frequency content of images (we use point-clouds with only 2048 points), or the fact that VGG was pre-trained while the PC-AE was not. 
However, we find \textit{significant} gains in accuracy ($4.1\%$ on average) from exploiting the two object representations {\em simultaneously}, implying a complementarity among them.

Next, we evaluate how the different approaches in incorporating context information described in Section~\ref{sec:neural_listeners} affect listener performance.
We focus on the more challenging object generalization task, using listeners that include attention and both object modalities. We report the findings in Table~\ref{table:listener_context_ablations}.
% \qq{}{Similar results on the language generalization task and details on the performance effect of using a convolutional instead an MLP layer, or non-permutation invariant ($f, g$) functions are given in the Appendix (Table \ref{multiple_listeners_language_task} and Section \ref{apndx:listener_details})}.
We find that the \listenerA \ and \listenerB \ models perform best overall, outperforming the \listenerC \ model, which does not share weights across objects.
%We find that the \listenerB \ architecture, which consumes the entire context with a single (non-weight shared replica) LSTM, performs significantly worse than both the \listenerA \ and \listenerC \ architectures (which use an explicit shared-weighting mechanism) and achieve similar performance to each other.
This pattern held for both hard and easy trial types in our dataset.
%It is plausible that our alternative strategies for incorporating context information however, would yield an advantage in hard contexts, where finer distinctions must be made. 
%However, we did not observe noticeable differences between the Separate and Separate-Augment variants in either the easy \emph{or} the hard subpopulations; we do find that easy contexts were easier for all models than hard contexts. 
We further explored the small portion (${\sim}14\%$) of our test set that use explicitly contrastive language: superlatives (``skinniest'')  and comparatives (``skinnier'').
Somewhat surprisingly we find that the \listenerA \ architecture remains competitive against the architectures with more explicit context information.
The \listenerA \ model thus achieves high performance and is the most flexible (at test time it can be applied to {\em arbitrary-sized} contexts); we focus on this architecture in the explorations below.

\subsection{Exploring learned representations}
\begin{figure}[tbh]
\includegraphics[scale=.5]{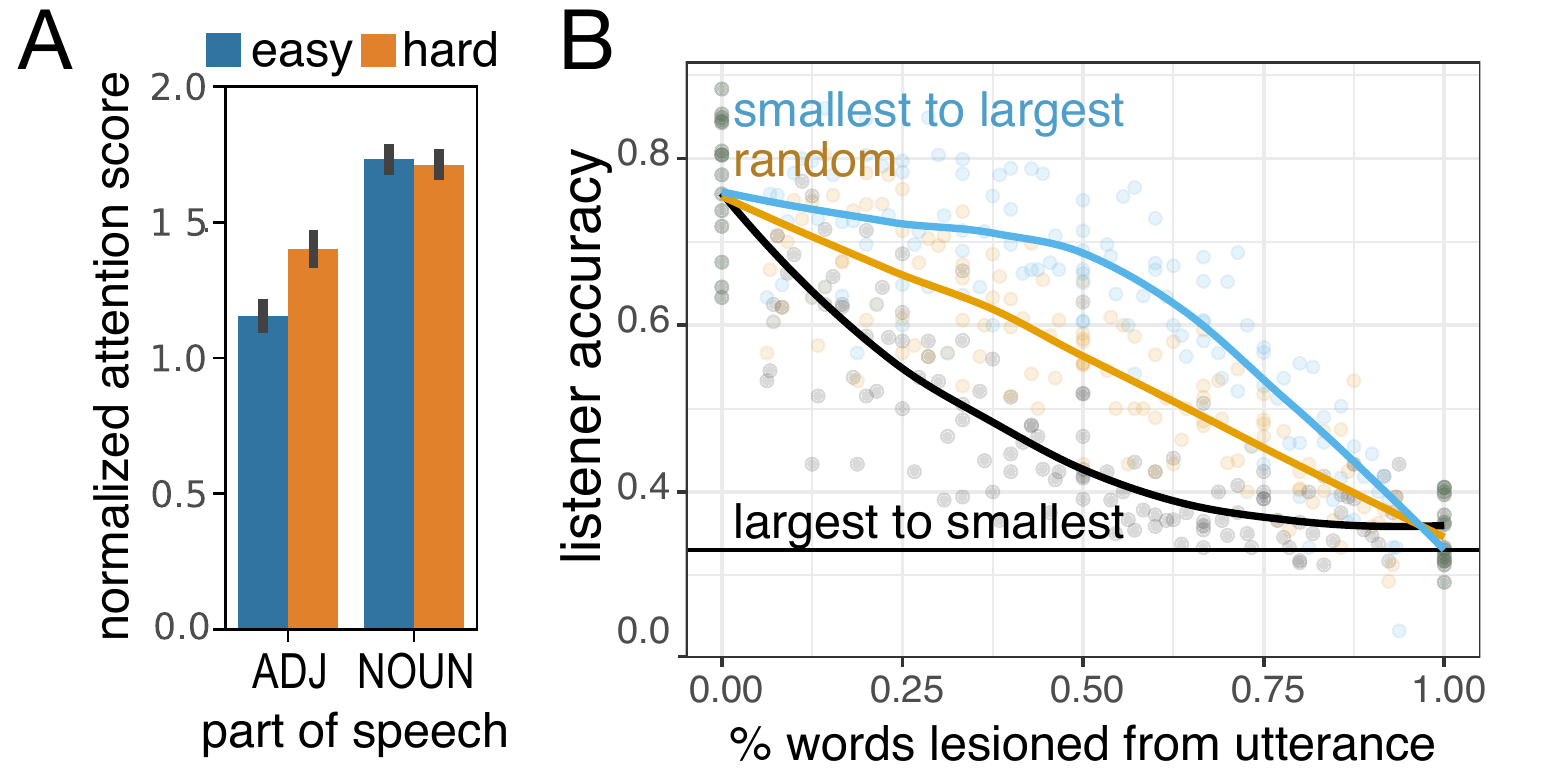}
\centering
\caption{(A) The listener places more attention on adjectives in hard (orange) triplets than easy (blue) ones. The histogram's heights depict mean attention scores normalized by the length of the underlying utterances; the error bars are bootstapped 95\% confidence intervals. (B) Lesioning highest attention words to lowest worsens performance more than lesioning random words or lesioning lowest attention words.}
\label{fig:word_attention}
\end{figure}

\paragraph{Linguistic ablations} Which aspects of a sentence are most critical for our listener's performance? 
To inspect the properties of words receiving the most attention, we ran a part-of-speech tagger on our corpus. 
We found that the highest attention weight is placed on \emph{nouns}, controlling for the length of the utterance. 
However, adjectives that \emph{modify} nouns received more attention in hard contexts (controlling for the average occurrence in each context), where nouns are often not sufficient to disambiguate (see Fig. \ref{fig:word_attention}A).
To more systematically evaluate the role of higher-attention tokens in listener performance, we conducted an utterance lesioning experiment. 
For each utterance in our dataset, we successively replaced words with the \texttt{<UNK>} token according to three schemes: (1) from highest attention to lowest, (2) from lowest attention to highest, and (3) in random order. 
We then fed these through an equivalent listener trained \emph{without} attention. 
We found that up to 50\% of words can be removed without much performance degradation, but only if these are low attention words (see Fig. \ref{fig:word_attention}B).
%This ablation result was found across a wide range of utterance lengths. 
Our word-attentive listener thus appears to rely on context-appropriate content words to successfully disambiguate the referent. 

\paragraph{Visual ablations} To test the extent to which our listener is relying on the same semantic {\em parts} of the object as humans, we next conducted a lesion experiment on the visual input. 
We took the subset of our test set where (1) all chairs had complete part annotations available \cite{yi2016_syncspeccnn} and (2) the corresponding utterance mentioned a {\em single} part (17\% of our test set). 
We then created lesioned versions of all three objects on each trial by removing pixels of images (and/or points when point-clouds are used), corresponding to parts according to two schemes: \emph{removing} a single part or \emph{keeping} a single part. We did this either for the mentioned one, or another part, chosen at random. We report listener accuracies on these lesioned objects in Table \ref{table:listener_part_lesion_img_only}.  We found that removing random parts hurts the accuracy by 10.4\% on average, but removing the mentioned part dropped accuracy more than three times as much, nearly to chance. Conversely, keeping  only the mentioned part while lesioning the rest of the image merely drops accuracy by 10.6\% while keeping a non-mentioned (random) part alone brings accuracy down close to chance. In other words, on trials when participants depended on information about a part to communicate the object to their partner, we found that visual information about that part was both {\em necessary and sufficient} for the performance of our listener model.

%!TEX root = ../sections/main.tex
% Non-attentive Listener, only images.
\begin{table}[tbh]
\centering
\begin{tabular}{p{7em}|p{5em}|p{5em}}
\hline
& 
\textbf{Single Part Lesioned} & \textbf{Single Part Present} \\ \hline
\textbf{Mentioned Part}       & $42.8\% \pm 2.3$               & $66.8\% \pm 1.4$ \\ \hline
\textbf{Random Part}          & $67.0\% \pm 2.9$               & $38.8\% \pm 2.0$ \\ \hline
\end{tabular}
\caption{Evaluating the part-awareness of neural listeners by lesioning object {\em parts}. 
Results shown are for image-only listeners, with average accuracy of $77.4\%$ when {\em intact} objects are used. Similar findings regarding point-cloud-based lesioning are provided in the Appendix.}
\label{table:listener_part_lesion_img_only}
\end{table}

\section{Neural speakers}
\label{Section:speaker_models}

\paragraph{Architecture} 
We next explore models that learn to generate an utterance that refers to the target and which distinguishes it from the distractors.
% Our baseline (\textit{literal}) speaker models are inspired by the show-and-tell model \cite{show-tell}. 
Similarly to a neural listener the heart of these models is an LSTM which encodes the objects of a communication context, and then decodes an utterance. Specifically, for an \textit{image-based} model, on the first three time steps, the LSTM input is the VGG code of each object. Correspondingly, for a \textit{point-cloud-based} model, the LSTM input is the object codes extracted from a PC-AE. During training and after the objects are encoded, the remaining input to the LSTM is the `current' utterance token, while the output of the LSTM is compared with the `next' utterance token, under the cross-entropy loss \cite{teacher_forcing}. 
% (After inputing the objects, this is a standard language model trained via teacher-forcing \cite{teacher_forcing}). 
The target object is always presented last, eliminating the need to represent the index of the target separately.
%During the first three time steps, the speaker receives sequentially the three object latent codes of a context (projected via an $L_2$-norm weight regularized FC) and outputs a vector which is transformed into a logit prediction over our vocabulary via an FC.
%The soft-normalized version of the output is compared against the first ground-truth token ($u_1$) under the cross-entropy loss. For each remaining token $u_k \in u_2, \ldots$, the LSTM is conditioned on the previous ($u_{k-1}$) ground-truth token and the cross-entropy comparison is repeated (i.e., we do teacher-forcing \cite{teacher_forcing}). In all speakers the target vector is fed third, thereby minimizing the length of dependence between the most important input object and the output~\cite{sutskever_seq_learning} and eliminating the need to represent the index of the target separately. 
To find the best model hyper-parameters (e.g.~$L_2$-weights, dropout-rate and \# of LSTM neurons) and the optimal stopping epoch, we sample synthetic utterances from the model during training and use a pretrained {\em listener} to select the result with the highest listener accuracy. We found this approach to produce models and parameters that yield better quality utterances than evaluating with listening-unaware metrics like BLEU \cite{BLEU}.

\paragraph{Variations}
The above (\textit{literal}) speakers can learn to generate language that discriminates targets from distractors.
To test the degree to which distractor objects are used for generation, we experiment with \textit{context-unaware} speakers that are provided the encoding of the target object {\em only}, and are otherwise identical to the above models. Motivated by the recursive social reasoning characteristic of human pragmatic language use (as formalized in the Rational Speech Act framework \cite{GoodmanFrank16_RSATiCS}), we create \textit{pragmatic} speakers that choose utterances according to their capacity to be discriminative, as judged by a pretrained ``internal" listener. In this case, we sample utterances from the (\textit{literal}) speakers, but score (i.e.~re-rank) them with:
\begin{equation}
%\beta \log(P_L(t|U)) + (1-\beta) \sum_{k=1}^{k=|U|} \frac{\log(P_S(u_k | O ))}{|U|^{\alpha}} \enspace ,
\beta \log(P_L(t|U,O)) + \frac{(1-\beta)}{|U|^{\alpha}}\log(P_S(U |O, t)),
\label{eq:speaker_scoring}
\end{equation}
where $P_L$ is the listener's probability to predict the target ($t$) and $P_S$ is the likelihood of the \textit{literal} speaker to generate $U$. 
The parameter $\alpha$ controls a length-penalty term to discourage short sentences \cite{Wu2016GooglesNM}, while $\beta$ controls the relative importance of the speaker's vs.\ the listener's opinions.
\section{Speaker experiments}

% Pragmatic vs. literal moved to dataset.tex

\begin{figure*}[tb]
\centering
\includegraphics[width=\textwidth]{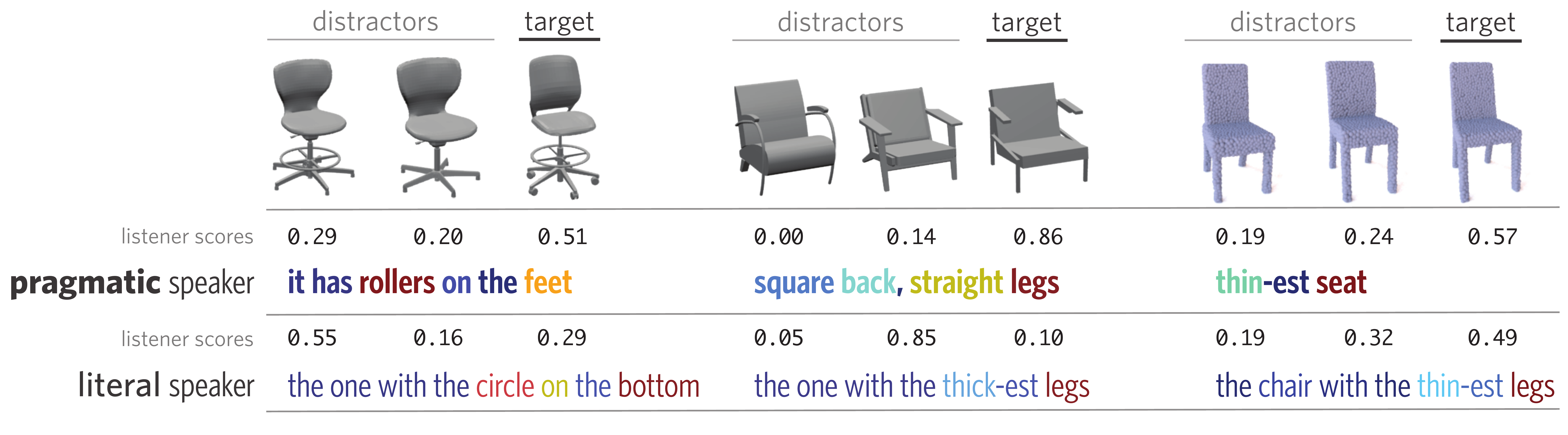}
\vspace{-20pt}
\caption{\textit{Pragmatic} vs. \textit{literal} speakers in {\em unseen} (`hard') contexts. The pragmatic generations successfully discern the target, even in cases where the literal ones fail. The two left-most examples are based on image-based speakers/listeners, the right-most with point-cloud-{\em based}. The utterances are color coded according to the attention placed by an evaluating neural listener whose classification scores are shown above each corresponding utterance.}
\label{fig:speaker_generations_close_cond_main}
\end{figure*}

\label{sec:speaker_results}
Qualitatively, our speakers produce good object descriptions, see Fig.~\ref{fig:speaker_generations_close_cond_main} for examples, with the pragmatic speakers yielding more discriminating utterances.\footnote{The project's webpage contains additional qualitative results.}
%Having established that our neural listener learns representations with good generalization, sensitive to task-relevant, (part-based) structure, we now proceed to evaluate our neural speakers.
% \footnote{Appendix Fig.~\ref{fig:context_effects} illustrates how the speakers refer to the same targets in far vs. close contexts. Appendix Figs
% \ref{fig:speaker_errors} and \ref{fig:listener_errors} showcase listener's and speaker's failure modes.} 
To quantitatively evaluate the speakers we measure their success in reference games with two different kinds of partners: with an independently-trained listener model and with human listeners. 
To conduct a {\em fair} study when we used a neural listener, we split the training data in half. The evaluating listener was trained using one half, while the scoring (or ``internal'') listener used by the pragmatic speaker was trained on the remaining half. For our human evaluation, we used the \textit{literal} and \textit{pragmatic} variants to generate referring expressions on the test set (we use all training data to train the internal listeners here). We then showed these referring expressions to participants recruited on AMT and asked them to select the object from context that the speaker was referring to. We collected approximately $2.2$ responses for each triplet (we use $1200$ unique triplets from the {\em object-generalization} test-split,
annotated separately by each speaker model). The synthetic utterances used were the highest scoring ones (Eq.~\ref{eq:speaker_scoring}) for each model with optimal (per-validation) $\alpha$ and a $\beta=1.0$. We note that while the \textit{point-based} speakers operate {\em solely} with point-cloud representations, we present their produced utterances to AMT participants accompanied by CAD rendered images, to keep the human-side presentation identical across experiments.

 % (More details including the effects of the parameters $\alpha$, $\beta$ and speaking in hard vs.~easy contexts are provided in the Supplemental Material).

% \begin{figure}[t!]
%         \centering
%         \includegraphics[scale=.22]{../figures/speaker_class_generalization.pdf}
%         \vspace{-5pt}
%         \caption{Our listener-aware speaker can produce informative referring expressions for {\em out-of-class} objects in context. Here, we apply our search technique in the collection of ShapeNet Tables, to produce triplets of well-separated objects. We use the queries: 'no legs' (left), 'modern' (center), and 'x' (right) to construct each context. Note how the target of each context reflects the semantics of the used query, as opposed to the distractors.}
%         \label{fig:speaking_out_of_class}
% \end{figure}

%!TEX root = ../sections/main.tex
\begin{table}[h!]
\centering
\caption{Evaluating neural speakers operating with Point Cloud or Image object representations.}
%When evaluating with neural listeners, five random seeds controlling the weight initialization and speaker-listener data splits were used.}
 \begin{tabular}{c| c| c| c} 
 \hline
 \begin{tabular}{@{}c@{}}\textbf{Speaker} \\ \textbf{Architecture}\end{tabular}  &
 \textbf{Modality} & 
 \begin{tabular}{@{}c@{}}\textbf{Neural} \\ \textbf{Listener}\end{tabular} &
 \begin{tabular}{@{}c@{}}\textbf{Human} \\ \textbf{Listener}\end{tabular} \\
 \hline
 
 \begin{tabular}{@{}c@{}} Context \\ Unaware \end{tabular} &  
 					\begin{tabular}{@{}c@{}} Point Cloud \\ Image \end{tabular} &
 					\begin{tabular}{@{}c@{}} $ 59.1 \pm 2.0\%$  \\ $64.0 \pm 1.7\%$ \end{tabular} &
 					\begin{tabular}{@{}c@{}} - \\ - \end{tabular} \\
 \hline
Literal &  
 					\begin{tabular}{@{}c@{}} Point Cloud \\ Image \end{tabular} &
 					\begin{tabular}{@{}c@{}} 71.5 $ \pm 1.3 \%$ \\ $76.6 \pm 1.0\%$ \end{tabular} &
 					\begin{tabular}{@{}c@{}} 66.2 \\ 68.3 \end{tabular} \\
 \hline			
Pragmatic &  
 					\begin{tabular}{@{}c@{}} Point Cloud \\ Image \end{tabular} &
 					\begin{tabular}{@{}c@{}}  90.3 $ \pm 1.3 \%$ \\ {\bf 92.2} $\pm 0.5\%$ \end{tabular} &
 					\begin{tabular}{@{}c@{}}  69.4 \\ {\bf 78.7} \end{tabular} \\ 			

 % Literal  ($\beta=0.0$)       &  $76.6 \pm 1.0\%$ & 68.3 \\
 % Listener-aware ($\beta=0.5$)	&  $85.9 \pm 0.4\%$ & - \\
 % Listener-aware ($\beta=1.0$)	&  $92.2 \pm 0.5\%$ & 78.7\\

 \hline
\end{tabular}
\label{speaker-accuracy-chairs}
\end{table}

We found (see  Table~\ref{speaker-accuracy-chairs}) that our \textit{pragmatic} speakers perform best with both synthetic and human partners.
While their success with the synthetic listener model may be unsurprising, given the architectural similarity of the internal listener and the evaluating listener, \emph{human} listeners were $10.4$ percentage points better at picking out the target on utterances produced by the \textit{pragmatic} vs.~\textit{literal} speaker for the best-performing (\textit{image-based}) variant.
We also found an asymmetry between the listening and speaking tasks: while context-unaware listeners achieved high performance, we found that context-unaware speakers fare significantly worse than context-aware ones. Last, we note that both literal and pragmatic speakers produce {\em succinct} descriptions (average sentence length $4.21$ vs. $4.97$) but the pragmatic speakers use a much richer vocabulary ($14\%$ more unique nouns and $33\%$ more unique adjectives, after controlling for average length discrepancy).

% As a final qualitative examination of our speakers' generalization ability, we ran an \emph{out-of-class} speaking experiment. We constructed well-separated contexts from the search results presented in Section \ref{Section:search}, taking as the target a random example among the top-5 highest-scoring retrievals (give a query) and choosing distractors randomly from among the lowest-5 scoring. Our best speaker model produced promising results (see Fig. \ref{fig:speaking_out_of_class} for examples on tables, and Fig. \ref{apndx_fig:out_of_class_searching} of the Appendix for examples with sofas and lamps.
\section{Out-of-distribution transfer learning} 
\label{sec:ood_experiments}
Language is abstract and compositional. These properties make language use generalizable to new situations (e.g.~using concrete language in novel scientific domains) and robust to low-level perceptual variation (e.g.~lighting). In our final set of experiments we examine the degree to which our neural listeners and speakers learn representations that are correspondingly {\em robust}: that capture associations between the visual and the linguistic domains permit generalization out of the training domain.

\vspace{15pt}
\begin{figure*}[tb]
\centering
\includegraphics[width=\textwidth]{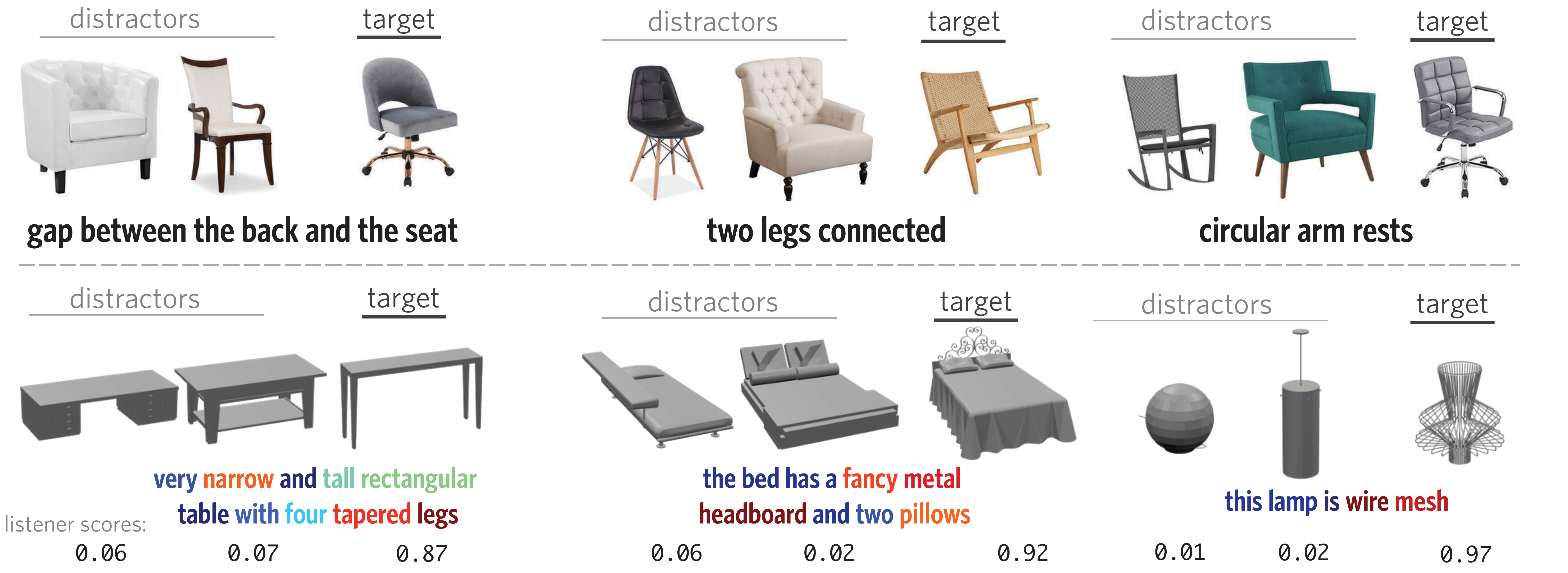}
\vspace{-10pt}
\caption{Examples of {\em out-of-distribution} neural speaking and listening. {\bf Top row}: model generations for \textit{real-world} catalogue images. The speaker successfully describes fine grained shape differences on images with rich color and texture content, not present in the training data. {\bf Bottom row}: results of applying a word-attentive listener on renderings of CAD objects from {\em unseen} classes with human-produced utterances. The listener can detect the (often localized) visual cues that humans refer to, despite the large visual discrepancy of these objects from training-domain of chairs. (The utterances are color coded according to the listener's attention.)}
\label{fig:out_of_class_pannel}
\end{figure*}

% \caption{Our neural-listeners are robust in predicting the referred object for a variety of unseen object classes. The examples shown include the human utterance describing the target with color-coded tokens, according to the attention a word-attentive listener assigns to each token (warmer is higher).}
% \label{fig:ood_mixed_panel}
% \vspace{-5pt}

% \begin{figure*}[t]
% \centering
% \includegraphics[scale=\textwidth]{../figures/out_of_class_mixed_pannel.pdf}
% \end{figure*}
% 
% \begin{figure}[t!]
% \centering
% \includegraphics[scale=0.20]{../figures/ood_rgb_neural_speaking.png}
% \caption{Out-of-distribution speaking. Our speaker learns fine-grained geometric properties of {\bf real chairs} with remarkable robustness to color and texture variations.}
% \label{fig:ood_rgb_neural_speaking}
% \vspace{-5pt}
% \end{figure}

\paragraph{Understanding out-of-class reference} To test the generalization of listeners to novel stimuli, we collected referring expressions in communication contexts made of objects in ShapeNet drawn from new classes: beds, lamps, sofas and tables. These classes are distinct from chairs, but share some parts and properties, making transfer possible for a sufficiently compositional model.
For each of these classes we created 200 contexts made of random triplets of objects; we collected 2 referring expressions for each target in each context (from participants on AMT). 
Examples of visual stimuli and collected utterances are shown in Fig.~\ref{fig:out_of_class_pannel} (bottom-row). 
To this data, we applied an (image-only, with/without-attention) listener trained on the CiC (i.e.~chairs) data. We avoid using point-clouds since unlike VGG which was finetuned with multiple ShapeNet classes, the PC-AE was pre-trained on a single-class.

As shown in Table~\ref{table:ood_human_listening_without_attn}, the average accuracy is well above chance in all transfer categories (56\% on average). 
Moreover, constraining the evaluation to utterances that contain {\em only} words that are in the CiC training vocabulary (75\% of all utterances, column: \textit{known}) only slightly improves the results. 
This is likely because utterances with unknown words still contain enough known vocabulary for the model to determine meaning.
We further dissect the \textit{known} population into utterances that contain part-related words (\textit{with-part}) and their complement (\textit{without-part}).
For the training domain of chairs without-part utterances yield slightly higher accuracy. However the useful subcategories that support this performance (e.g.~``recliner'') do not support transfer to new categories.
Indeed, we observe that for transfer classes (except sofa) the listener performs better when part-related words are present. 
Furthermore, the performance gap between the two populations appears to become larger as the perceptual distance between the transfer and training domains increases (compare sofas to lamps).

%!TEX root = ../sections/main.tex
\begin{table}[tbh]
\caption{Transfer-learning of neural listeners in novel object {\em classes}, and in different subpopulations of utterances. The subpopulations are: \textit{entire}: all utterances, \textit{known}: with all tokens in the chair training-vocabulary, \textit{with-part}: subset of \textit{known} that contain at least one part-related word, \textit{without-part} subset of \textit{known} and complement of \textit{with-part}. For reference the test-chair statistics are shown (first row), but are not included in the reported average (last row). The accuracies are {\em averages} of five listeners trained on different data splits. Further details are provided in the Appendix.}
\label{table:ood_human_listening_without_attn}
\begin{tabular}{|l||r|r|r|r|}
\hline
& \multicolumn{4}{|c|}{Population} \\
\hline
Class    &  entire & known & with part & without part\\
\toprule
chair    & 77.4 & 77.8  &  77.0       &  {\bf 80.5}  \\
\midrule
bed      & 56.4 & 55.8  & {\bf 63.8}  &  51.5   \\
\hline
lamp     & 50.1 & 51.9  & {\bf 60.3}  &  47.1   \\
\hline
sofa     & 53.6 & 55.0  & {\bf 55.1}  &  54.7   \\
\hline
table    & 63.7 & 65.5  & {\bf 68.3}  &  62.7    \\
\hline
average  & 56.0 & 57.1  & {\bf 61.9}  &  54.9    \\
\hline
\end{tabular}
\end{table}
\vspace{-15pt}

\paragraph{Describing real images}
Transfer from synthetic data to real data is often difficult for modern machine learning models, that are attuned to subtle statistics of the data. 
We explored the ability of our models to transfer to real chair images (rather than the training images which were rendered without color or texture from CAD models) by curating a modest-sized (300) collection of chair images from online furniture catalogs. These images were taken from a {\em similar} view-point to that of the training renderings and have rich color and texture content. 
We applied the (image-only) \textit{pragmatic} speaker to these images, after subtracting the average ImageNet RGB values (i.e.~before passing the images to VGG). 
Examples of the speaker's productions are shown in Figure~\ref{fig:out_of_class_pannel}. 
For each chair, we randomly selected two distractors and asked 2 AMT participants to guess the target given the utterance produced by our speaker.  
Human listeners correctly guessed the target chair $70.1\%$ of the time.
Our speaker appears to transfer successfully to real images, which contain color, texture, pose variation, and likely other differences from our training data.

\section{Related work}
\label{sec:rel_work}

\textbf{Image labeling and captioning} \hspace{3mm}
Our work builds on recent progress in the development of vision models that involve some amount of language data, including object categorization \cite{simonyan2014very,zhang2014part} and image captioning \cite{karpathy2015deep, show-tell,show_attend_tell}. Unlike object categorization, which pre-specifies a fixed set of class labels to which all images must project, our systems use open-ended, referential language. Similarly to other recent works in image captioning \cite{mao16,nagaraja16, licheng_16, unsup_context_aware, luo2017comprehension, baby_talk, mattnet}, instead of captioning a single image (or entity therein), in isolation, our systems learn how to communicate across diverse communications contexts.

\textbf{Reference games} \hspace{3mm}
In our work we use reference games \cite{referit} in order to operationalize the demand to be relevant in context. The basic arrangement of such games can be traced back to the language games explored by Wittgenstein \cite{wittgenstein1953philosophical} and Lewis \cite{Lewis69_Convention}. For decades, such games have been a valuable tool in cognitive science to quantitatively measure inferences about language use and the behavioral consequences of those inferences \cite{rosenberg1964speakers,KraussWeinheimer64_ReferencePhrases, ClarkWilkesGibbs86_ReferringCollaborative, VanDeemter16_ComputationalModelsOfReferring}. Recently, these approaches have also been adopted as a benchmark for discriminative or context-aware NLP \cite{PaetzelRaccaDeVault14_RDGCorpus, andreas2016reasoning, su2017reasoning, vedantam2017context, monroe_colors, cohn2018pragmatically, lazaridou2018emergence}.

\textbf{Rational speech acts framework} \hspace{3mm}
Our models draw on recent formalization of human language use in the Rational Speech Acts (RSA) framework \cite{GoodmanFrank16_RSATiCS}.
At the core of RSA is the Gricean proposal \cite{Grice75_LogicConversation} that speakers are agents who select utterances that are parsimonious yet informative about the state of the world. RSA formalizes this notion of informativity as the expected reduction in the uncertainty of an (internally simulated) listener, as our pragmatic speaker does.
The literal listener in RSA uses semantics that measure compatibility between an utterance and a situation, as our baseline listener does. Previous work has shown that RSA models account for context sensitivity in speakers and listeners \cite{GrafEtAl16_BasicLevel, monroe_colors, speaker_listener_reinforce, unified_pragmatics}. Our results add evidence for the effectiveness of this approach in complex domains.
\section{Conclusion}
In this paper, we explored models of natural language grounded in the shape of common objects.
The geometry and topology of objects can be complex and the language we have for referring to them is correspondingly abstract and compositional. This makes the shape of objects an ideal domain for exploring grounded language learning, while making language an especially intriguing source of evidence for shape variations. We introduced the Chairs-in-Context corpus of highly descriptive referring expressions for shapes in context. Using this data we explored a variety of neural listener and speaker models, finding that the best variants exhibited strong performance.
These models draw on both 2D and 3D object representations and appear to reflect human-like part decomposition, though they were never explicitly trained with object parts.
Finally, we found that the learned models are surprisingly robust, transferring to real images and to new classes of objects.
Future work will be required to understand the transfer abilities of these models and how this depends on the compositional structure they have learned.
% \section*{Acknowledgements}
\paragraph{Acknowledgements}
The authors wish to acknowledge the support of a Sony Stanford Graduate Fellowship, a NSF grant CHS-1528025, a Vannevar Bush Faculty Fellowship and gifts from Autodesk and Amazon Web Services for Machine Learning Research.

%%%%%%%%% REFERENCES
{\small
\bibliographystyle{aux/ieee}
\bibliography{aux/references}
}

\clearpage
%%%%%%%%% APPENDIX
\appendix
\section{Appendix}

\begin{figure*}[tb]
	\centering
	\includegraphics[width=\textwidth]{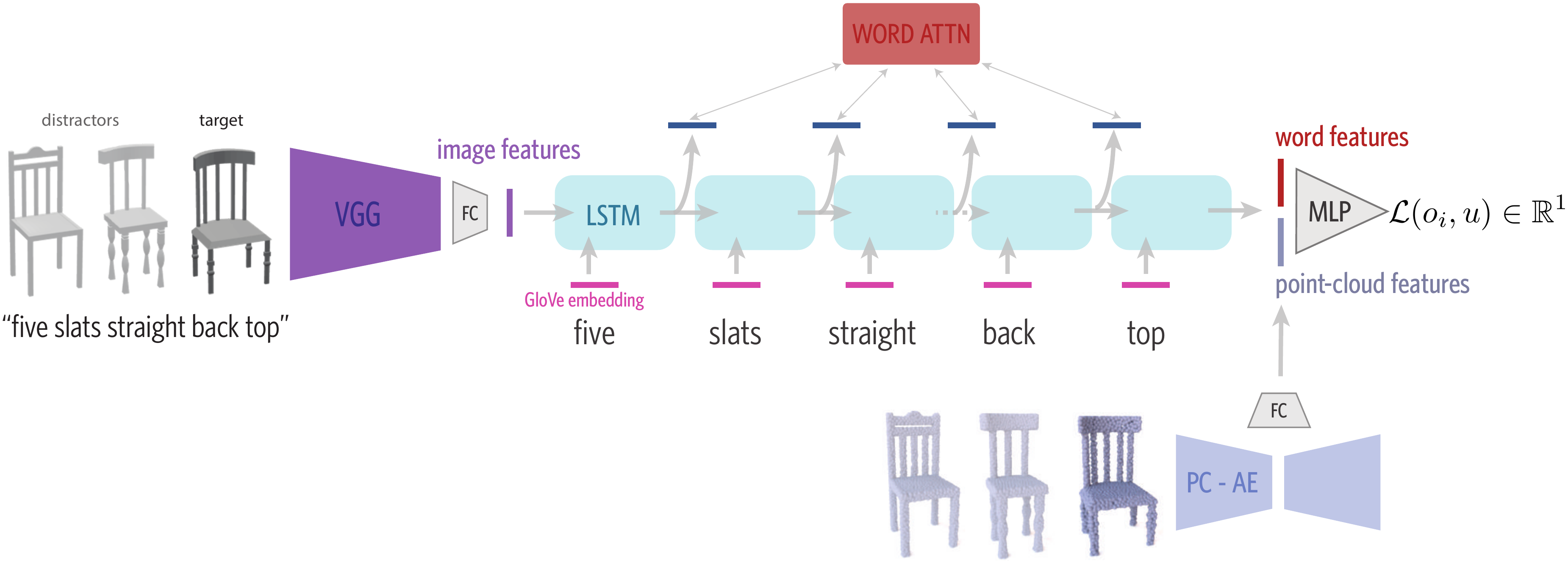}
	\caption{\listenerA\ listener architecture combining 2D images, 3D point-clouds and linguistic utterances.}
	\label{proposed_listener_arch}
\end{figure*}

\subsection{\textbf\textit{CiC} details}
\label{apndx:building_triplets_details}
To build the triplets comprising the communication contexts of \cic, we exploited the \textit{latent} (bottleneck-derived) vector space of a Point-Cloud based AutoEncoder (PC-AE) \cite{achlioptas2018latent_pc}, trained with chair-only objects of ShapeNet \cite{shapenet}. Concretely, we used a PC-AE with small bottleneck (64D) to promote meaningful euclidean distances and after embedding all $\sim 7000$ ShapeNet chairs in the resulting space, we computed their underlying 2-(euclidean)-nearest-neighbor graph. On this graph, we selected the $1K$ chairs with the highest in-degree to `seed' the triplet generation. For each of the 1K (seed) chairs, we considered it together with its two nearest neighbors from the {\em entire} shape collection to form a \textit{Hard} triplet. Also, we considered it together with the two chairs that were closest to it but which were also more distant from it than the median of all pairwise distances, to form an \textit{Easy} triplet.
The above procedure gives rise to 2000 communication contexts when target vs. distractor information is ignored. However, to counterbalance the dataset while annotating these contexts in AMT, we ensured that {\em each} chair of a context was considered as a distractor and as a target, and that each resulting combination was annotated by at least 4 humans. Last, we note that when building the Hard triplets, we applied a manually tuned distance-threshold, to reject triplets that contained objects that were `too' close: we found that about $\sim 3\%$ of chairs had a geometric duplicate that could vary only wrt. its texture.

\subsection{Image and point-cloud pre-training}
For the listeners and speakers we trained a PC-AE under the Chamfer loss~\cite{achlioptas2018latent_pc} with a 128D bottleneck and point clouds with 2048 points extracted from 3D CAD models, uniformly area-wise.  We also fine-tuned a VGG-16 pre-trained on ImageNet on a 8-way classification, with 36,632 rendered images of textureless 3D CAD models, taken from a single view-point. Concretely, we used images of the 8 largest object classes of Shape-Net (car, airplane, vessel, sofa, chair, table, lamp, riffle) and a uniformly random i.i.d. split of [90\%, 5\%, 5\%] for train/test/val purposes. We fine-tuned the network for 30 epochs. During the first 15 epochs we optimized \textit{only} the weights of the last (fc8) layer and during the last 15 epochs the weights of all layers. The attained test classification accuracy was $96.9\%$. Last, to embed an image for the downstream listening/speaking tasks, we used the 4096D output activations of the penultimate (fc7) fully-connected layer. 

\subsection{Pre-processing utterances}
We preprocessed the collected human utterances by i) lowercasing, ii) tokenizing by splitting off punctuation, iii) tokenizing by splitting superlative or comparative adjectives ending in -er, -est to their stem word, e.g.~`thinner:' $\rightarrow$ [`thin', 'er'] and, iv) replacing tokens that appear once or not at all in a training split with a special symbol marking an unknown token (\texttt{<UNK>}). Furthermore, we ignored the utterances comprised by more than 33 tokens (99th percentile) and those for which the human listener in the underlying trial did not guess correctly the target.  Last, we concatenated listener and speaker utterances from the same trial (in their order of formulation) by adding in the end of each but the last utterance a special symbol marking a dialogue: (\texttt{<DIA>}), e.g.~[`the', `thin', `chair', \texttt{<DIA>}, `yes'].

%!TEX root = ../../sections/main.tex

\begin{table*}[tbh!]
 \centering
 \begin{tabular}{|c||c|c|c|} 
 \hline
 \backslashbox{Hyper\\Parameters}{\begin{tabular}{@{}c@{}}Architecture\end{tabular}}& \listenerA & \listenerB & \listenerC\\

 \hline
 Learning rate         & 0.0005 &  0.001 &  0.001\\ 
 Label-smoothing       & 0.9    &  0.9   &  0.9\\
 $L_2$ regularization  & 0.3    &  0.05  &  0.09\\
 LSTM-input-dropout    & 0.5    &  0.7   &  0.45\\
 \hline
 
\end{tabular}
\caption{Optimal hyper-parameters for ablated neural listener architectures, using both geometric modalities and word-attention and various degrees of context. Dropout numbers reflect the \textit{keep} probability.}
\label{table:listener-optimal-hyper-params}
\end{table*}

\subsection{Listeners details}
\label{apndx:listener_details}
For the listeners we used a uni-directional LSTM cell with $100$ hidden units, the output of which was passed into a 3-layer MLP with [100, 50, 3] neurons that predicted the triplet's classification logits. To the output of each hidden layer of the MLP, batch normalization \cite{bnorm} and a ReLU \cite{maas2013_rectifier-nonlinearities-improve-neural} non-linearity was applied.
The listeners' word-embedding was initialized with a $100D$ GloVe embedding pre-trained on the 6B Wikipedia 2014 corpus, and which was further fine-tuned during training. The PC-AE (128D) and VGG (4096D) latent vectors, that encoded each object, were passed as {\em input} to the LSTM when only one geometric modality was used. When the two modalities used together, the PC-AE codes were concatenated with the {\em output} of the LSTM, and the concatenated result was processed by the final MLP. In either case, we first re-embedded these geometric codes (100D) with 2 separate/single FC-ReLU layers (referred as `projection' layers in the Main Paper Section~\ref{sec:neural_listeners}). An overview of the proposed listener reflecting the overall design choices is given in Fig.\ref{proposed_listener_arch}. We used dropout with 0.5 keep probability {\em before} the `projection' layers with a drop-out mask that was the same for the objects of a given triplet.  Separate dropout with 0.5 keep probability was applied in all input vectors of the LSTM (i.e.~ on the language tokens or the grounding geometric codes). Last, the ground-truth indicator vectors of each triplet were label-smoothed \cite{rethinking_inception} by assigning 0.933 probability mass to the target and 0.0333 to the distractors (i.e.~smoothing of 0.9).

\paragraph{Discussion}
Label smoothing yielded a mild performance boost of $\sim 2\%$ across all ablated listener architectures, in accordance with previous work \cite{rethinking_inception}. We note that we did not manage to improve the best attained accuracies by applying layer normalization \cite{rnn_layer_norm} in the LSTM, or adversarial regularization \cite{Adv_noise} on the word-embedding. Dropout \cite{srivastava2014_dropout:-a-simple-way-to-prevent-neural} was by far the most effective form of regularization for our listeners ($\sim$[8-9]\%), following by $L_2$ weight-regularization of the projection layers ($\sim$[2-3]\%).
Finally, using a separate MLP to process the PC-AE codes, was slightly better than feeding them directly in the LSTM (after the tokens of each utterance were processed). However, grounding the LSTM with the PC-AE codes, and using the VGG codes in the end of the pipeline (either via pre-MLP concatenation or by feeding the latter in the LSTM) deteriorate {\em significantly} all attained results.

\paragraph{Context Ablations} We ablated three architectures that used simultaneously images and point-clouds, word attention and different degrees of context (See Main Paper Section~\ref{sec:neural_listeners}). The optimal Hyper-Parameters (HP) for each architecture are shown in Table~\ref{table:listener-optimal-hyper-params}. We did a grid search over the space of HP associated with each architecture {\em separately}. To circumvent the exponential growth of this space, we search it into two phases. First, we optimized the learning rate (in the regime of [0.0001, 0.0005, 0.001, 0.002, 0.004, 0.005])
in conjunction with the drop-out (keep probability) applied at the LSTM's \textit{input}, in the range [0.4-0.7] with increments of 0.05. Given the acquired optimal values, we searched for the optimal $L_2$ weight-regularization (in the range of [0.005, 0.01, 0.05, 0.1, 0.3, 0.9]) applied at the two projection layers, and label-smoothing ([0.8, 0.9, 1.0]). For these experiments we used a single random seed to control for the data splits with the \textit{object-generalization task}. We note that for the \listenerB \ listener, using a single 1D convolutional layer to extract the grounding vector of each object, appeared to produce better results than using a single FC layer (or deeper alternatives). This single convolutional layer we used, converted the input signal $\lbrack f(v_j, v_{k}) || g(v_j, v_{k}) || v_i \rbrack$ $\in \reals^{100 \times 3}$ 
to a $\reals^{100 \times 1}$ LSTM-grounding vector for {\textit each} object $v_i$, with an $8 \times 3 \times 1$ kernel and stride $1$.
 % (See Main Paper Section~\ref{sec:neural_listeners} for details on the input signal).

\paragraph{Training} We trained the \listenerA \ and the \listenerC \ for $500$ epochs and the \listenerB \ for $350$. This was sufficient, as more training increased overfitting without improving the attained test/val accuracies. We halved the learning every 50 epochs, if the validation error was not improved in any of them. Namely, every 5 epochs we evaluated the model on the validation split in order to select the epoch/weights with the best accuracy. Because the \listenerC \ is sensitive in the input order of the object codes, we randomly permute them during training. We use the ADAM \cite{ADAM} ($\beta_1 = 0.9$) optimizer for all experiments.

\subsection{Speaker details}
\label{apndx:speaker_details}
\paragraph{Image-based speaker}
To find good model parameters for an image-based speaker, we considered a hyper-parameter search on a {\em literal} variant. Similarly, to what we did in the ablations of listener variants we conducted a two-stage grid search given a single random seed and the \textit{object generalization} task. At the first stage, we searched models varying: a) the hidden neurons of the LSTM ($100$ or $200$), b) the initial learning rate ([0.0005, 0.001, 0.003]), c) the drop-out keep probability applied on the word-embeddings ([0.8, 0.9, 1.0]) and d) the dropout keep probability applied at the LSTM's output ([0.8, 0.9, 1.0]). The two best performing models were further optimized by considering $L_2$-weight regularization applied at the FC-projection layer (with values in [0, 0.005, 0.01]) and the dropout keep-probability applied before the FC-projection layer ([0.5, 0.7, 0.9 1.0]). The resulting optimal parameters are reported in Table~\ref{table:speaker-optimal-hyper-params}.
%!TEX root = ../../sections/main.tex

\begin{table*}[tbh!]
\centering
 \begin{tabular}{|c|c|c|c|c|c|} 
 \hline
 LSTM Size & Learning rate & $L_2$-reg. & Word-Dropout & Image-Dropout  & LSTM-out Dropout\\
 \hline
 200  & 0.003 &  0.005 &  0.8 & 0.5 & 0.9\\ 
 \hline
\end{tabular}
\caption{Optimal hyper parameters for \textit{literal} image-based neural-speaker. The dropout numbers reflect keep probabilities and the Image-Dropout refers to the dropout applied at the VGG-image codes, before the FC-projection layer.}
\label{table:speaker-optimal-hyper-params}
\end{table*}

\paragraph{Point-cloud-based speaker}
For the point-based speaker, we did a similar but more constrained hyper-parameter search as we did for the image-based speaker, by also considering its {\em literal} variant. Here, we fixed the drop-out applied the word-embeddings and to 
the LSTM's output (0.8 and 0.9 keep-probability respectively) and ablated the remaining hyper-parameters as we did for the image-based speaker. We found the same configuration of parameters (Table~\ref{table:speaker-optimal-hyper-params}) to be optimal for point-based models as well. Exception to this was the the dropout applied to the PC-AE codes before the FC-projection (no dropout at all was best in this case). Also, the point-based speakers needed more training to converge than the image-based ones (maximally 400 epochs vs. 300).

\paragraph{Model selection} To do model selection for a training speaker, we used a pre-trained listener (with the same train/test/val splits) which evaluated the synthetic utterances produced by the speaker during training. To this purpose the speaker generated 1 utterance for each unique triplet in the validation set via greedy (arg-max) sampling every 10 epochs of training and the listener reported the accuracy of predicting the target given the synthetic utterance. In the end of training (300 epochs for image-based speakers vs. 400 for point-based ones), the epoch/model with the highest accuracy was selected.

\paragraph{Other details}
We initially used GloVe to provide our speakers pre-trained word embeddings, as in the listener, but found that it was sufficient to train the word embedding from uniformly random initialized weights (we used the range [-0.1, 0.1]). 
We also initialized the bias terms of the linear word-encoding layer with the log probability of the frequency of each word in the training data \cite{karpathy2015deep}, which provided faster convergence. We train with SGD and ADAM ($\beta_1 = 0.9$) and apply norm-wise gradient clipping with a cut-off threshold of 5.0. The training utterances have a maximal length of 33 tokens (99th percentile of the dataset). For any speaker we sampled utterances of the maximum training length. For the {\em pragmatic}
speaker we sample and score $50$ utterances per triplet at test time (following Eq.~\ref{eq:speaker_scoring} of Main Paper).

\paragraph{Point-cloud \& image-based speaker}

%!TEX root = ../../sections/main.tex
\begin{table}[h]
\centering
 \begin{tabular}{||c||c||} 
 \hline
 Approach & Listener's Accuracy \\
 \hline
 Concat (100D) & $65.1 \pm 0.51\%$\\ 
 Concat (200D) & $78.2 \pm 0.95\%$\\
 Sum           & $77.9 \pm 0.38\%$ \\
 Serial        & ${\bf 79.0} \pm 0.32\%$ \\
 \hline
\end{tabular}
\caption{Ablating approaches for incorporating simultaneously point-clouds with images in a \textit{literal} neural-speakers. \textit{Sum}: Summing the two latent codes for each object. \textit{Concat}: Concatenating the codes. \textit{Serial}: Feeding them one after the other in the LSTM. Concatenation naturally doubles the input-dimensions of the LSTM (Concat 200D). To keep them the same as with all other experiments (100D) we also tested reducing the VGG/PC-AE projection layers to 50 dimensions for each modality (Concat 100D). Results are averages of 5 samples of utterances for a fixed test dataset.}
\label{table:ablation-speaker-images-and-pcs}
\end{table}

In \textit{preliminary} experiments, we attempted to incorporate both geometric modalities: point-clouds and images in a speaker network, similarly to what we did for the best-performing listener. While, this resulted in a (\textit{literal}) speaker model that could achieve higher neural-listener evaluation-accuracy than when either modality was used in isolation, we did not observe any improvement against the image-based speaker in AMT \textit{human}-listener experiments.

We attempted three ways of `mixing' the two modalities in a speaker. Namely, for each object of a communication context:  a) providing the LSTM with the \textit{concatenation} of its projected VGG code and its projected PC-AE code, b) same as a) but instead of concatenation, using the \textit{sum} operator, c) first providing its PC-AE projected code followed at the \textit{next time} step by its VGG one. We compared these approaches by using the optimal hyper-parameters for an image-based speaker and only vary the amount of dropout applied to the point-cloud before the projection layer ([1.0 0.8, 0.6] keep probability). In all cases, avoiding dropout was best. The final results for a single random-seed and the object-generalization task are reported in Table~\ref{table:ablation-speaker-images-and-pcs}. We note that while the optimal speaker that used two modalities performed slightly better than the image-based speaker, per neural-listener evaluation, it did not improve the attained performance in preliminary experiments with of human listeners in AMT.

% \newpage
\subsection{Further quantitative results}

\subsubsection{Listeners: context incorporation}

%!TEX root = ../../sections/main.tex
\begin{table*}[t!]
\centering
\begin{tabular}{c| c | c c c c c}
\hline
\multirow{2}{*}{\textbf{Architecture}} & 
	& \multicolumn{5}{c}{\textbf{Subpopulations}}  \\
	& \textbf{Overall} & \textbf{Hard} & \textbf{Easy} & \textbf{Sup-Comp} & \textbf{Negative} & \textbf{Split} \\
\hline

\listenerC
&$75.9 \pm 0.5\%$
&$67.4 \pm 1.0\%$
&$83.8 \pm 0.6\%$
&$74.4 \pm 1.5\%$
&$77.3 \pm 1.5\%$
&$65.8 \pm 5.2\%$\\

\listenerB
&$79.4 \pm 0.8\%$
&$\textbf{70.1} \pm 1.3\%$
&$88.1 \pm 0.6\%$
&$75.6 \pm 2.2\%$
&$\textbf{78.9} \pm 1.4\%$
& $\textbf{67.4} \pm 3.6\%$\\

\listenerA
&$\textbf{79.6} \pm 0.8\%$
&$69.9 \pm 1.3\%$
&$\textbf{88.8} \pm 0.4\%$
&$\textbf{76.3} \pm 1.3\%$    
&$77.5 \pm 1.2\%$
&$62.5 \pm 3.7\%$\\
\hline
\end{tabular}
\caption{
Comparing the effect of context inspection for listening on various (test) subpopulations of the \textit{object generalization} task. The listeners use images, point-clouds and word-attention. Reporting averages of five random seeds controlling the split populations and the network's initialization.}
\label{table:multiple_listeners_object_task}
\end{table*}

%!TEX root = ../../sections/main.tex
\begin{table*}[t!]
\centering
\begin{tabular}{c| c | c c c c c}

\hline
\multirow{2}{*}{\textbf{Architecture}} & 
	& \multicolumn{5}{c}{\textbf{Subpopulations}}  \\
	& \textbf{Overall} & \textbf{Hard} & \textbf{Easy} & \textbf{Sup-Comp} & \textbf{Negative} & \textbf{Split} \\
\hline

\listenerC
&$78.4 \pm 0.2\%$ 
&$71.5 \pm 0.6\%$
&$85.2 \pm 0.3\%$
&$75.8 \pm 0.9\%$
&$77.6 \pm 0.8\%$
&$61.8 \pm 3.0\%$\\

\listenerB
&$\textbf{84.4} \pm 0.5\%$ 
&$\textbf{78.5} \pm 0.8\%$ 
&$90.2 \pm 0.7\%$          
&$\textbf{80.9} \pm 0.6\%$ 
&$\textbf{82.6} \pm 1.1\%$
&$\textbf{68.9} \pm 2.3\%$\\

\listenerA
&$83.7 \pm 0.2\%$          
&$77.0 \pm 0.8\%$         
&$\textbf{90.3} \pm 0.3\%$ 
&$80.8 \pm 0.8\%$
&$80.5 \pm 1.0\%$
&$64.6 \pm 3.7\%$ \\

\hline
\end{tabular}
\caption{Comparing the effect of context inspection for listening on various (test) subpopulations of the \textit{language generalization} task. The listeners use images, point-clouds and word-attention. Reporting averages of five random seeds controlling the split populations and the network's initialization.}
\label{table:multiple_listeners_lang_task}
\end{table*}

In Table~\ref{table:multiple_listeners_object_task} we complement the results presented in the Main Paper at Table~\ref{table:listener_context_ablations}, by including two more sub-populations ('Negative' and 'Split'). In Table~\ref{table:multiple_listeners_lang_task}, we repeat this study for listeners trained and tested on the \textit{language generalization} task. 'Negative' is a subpopulation of utterances that contain at least one word of negative content e.g. 'not', 'but' etc. and is comprised by $\sim 15.0\%$ of all test utterances. 'Split' is smaller subpopulation ($\sim3.2\%$ of test data) that includes language the explicitly contrasts the target with the distractors e.g.~ `from the two that have thin legs, the one...'. We used an ad hoc set of search queries to find such utterances among the test set and found that the \listenerB\ architecture does perform noticeably better on these utterances. However, given the low occurrence of such cases, the resulting effects were not significant and we decided the gains of \listenerB\ architecture were not worth the increase in model complexity and rigidity with respect to context size.

\subsubsection{Listeners: part-lesion}
%!TEX root = ../../sections/main.tex

% Non-attentive Listener, images + pc.
\begin{table}[htp]
\centering
\begin{tabular}{p{7em}|p{5em}|p{5em}}
\hline
& 
\textbf{Single Part Lesioned} & \textbf{Single Part Present} \\ \hline
\textbf{Mentioned Part}       & $44.9\% \pm 1.2$               & $67.2\% \pm 1.1$ \\ \hline
\textbf{Random Part}          & $68.9\% \pm 1.3$               & $42.3\% \pm 1.3$ \\ \hline
\end{tabular}
\caption{Evaluating the part-awareness of neural listeners by lesioning object {\em parts}. 
Results shown are for listeners using {\bf both} point-clouds and images, with average accuracy of $78.8\%$ when {\em intact} objects are used.}
\label{table:listener_part_lesion_img_and_pc}
\end{table}

% CORRECTED caption: The performance of intact objects is 77.8
We complement Table~\ref{table:listener_part_lesion_img_only} of the Main Paper, with a similar study (Table \ref{table:listener_part_lesion_img_and_pc}) where we ablate our neural listeners with regards to their sensitivity in referential utterances based on object parts, when \textit{both} geometric modalities are used. We have observed that the PC-AE attempts to reconstruct (decode) noisy but {\textit complete} models, even when the input is a partial, which could explain the gains seen in Table~\ref{table:listener_part_lesion_img_and_pc} compared to Table~\ref{table:listener_part_lesion_img_only} when lesioning parts.

%!TEX root = ../../sections/main.tex
\begin{table}
\centering
\begin{tabular}{|l||r||r|r|}
\hline
& \multicolumn{3}{|c|}{Population} \\
\hline
Class    &  entire & with part & without part \\
\toprule
chair    & 7.1     & 8.0 (77\%)  & 4.7 (21\%) \\
\midrule
bed      & 6.4     & 7.0 (26\%)  & 5.3 (48\%) \\
\hline
lamp     & 7.3     & 11.0 (20\%) & 5.9  (37\%)\\
\hline
sofa     & 10.1    & 11.0 (72\%) & 5.9  (15\%)\\
\hline
table    & 6.6     & 8.0 (40\%)  & 4.9 (42\%) \\
\hline
average  & 7.6     & 9.3 (39.5\%) &  5.5 (35.5\%) \\
\hline
\end{tabular}
\caption{Average length of utterances for various transfer classes (complementing Table~\ref{table:ood_human_listening_without_attn}, Main Paper).  Between parentheses is reported the percentage of the entire population that is captured by its specific sub-population. The average (last row) is wrt. the transfer classes only; the chair-category is displayed for reference.}
\label{table:ood_human_listening_basic_data_statistics}
\end{table}

\subsubsection{Speakers: length penalty and listener awareness}
\label{alpha_beta_ablations}
%%% Note, image-based speakers are used for these ablation, with listeners that use point-clouds + images + word-attention.

\begin{figure*}[h]
\centering
	\begin{subfigure}{.5\textwidth}
	\centering
	\includegraphics[scale=0.45]{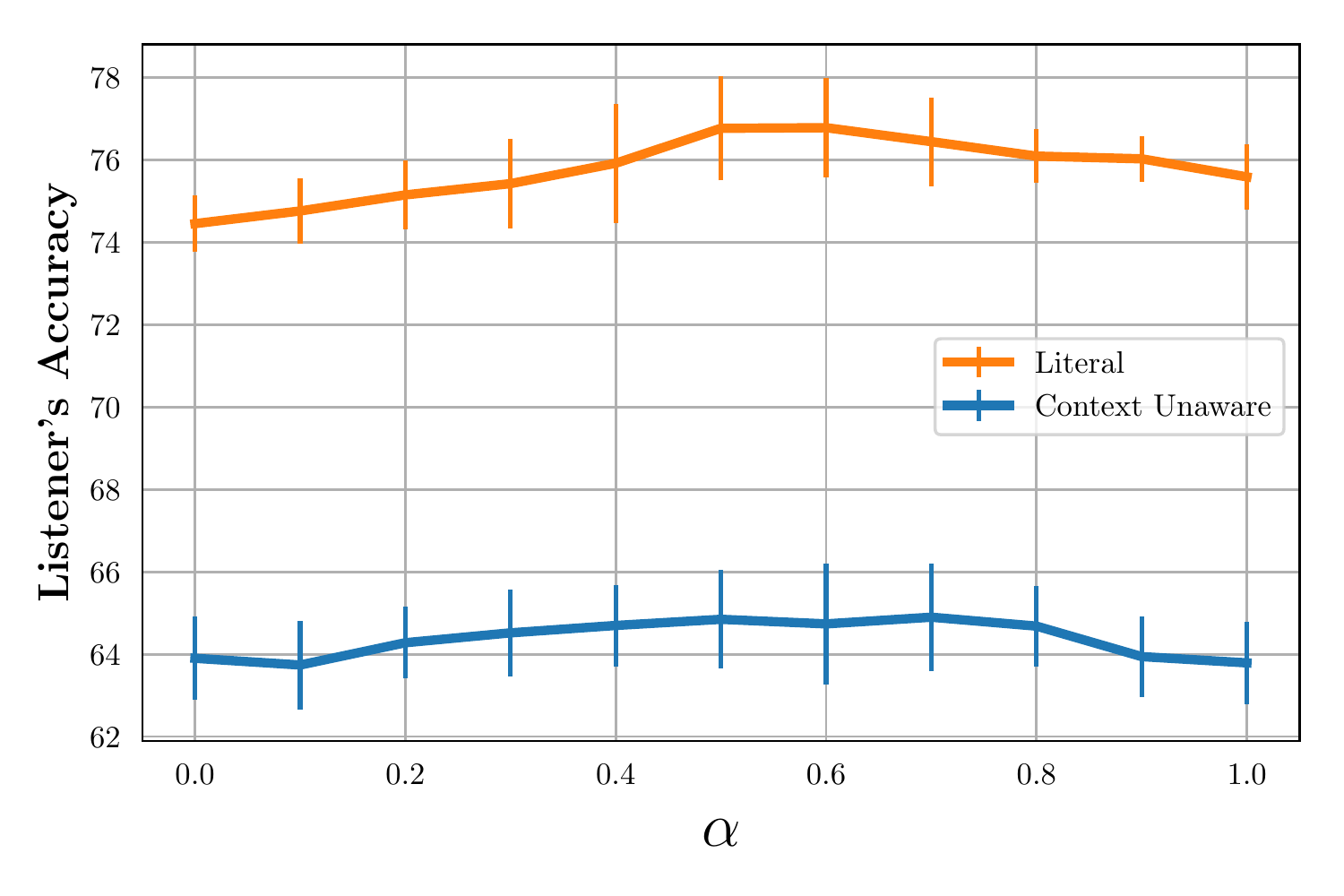}
	\caption{Effect of the length-penalty.}
	\label{fig:alpha_ablations}
	\end{subfigure}%
	\begin{subfigure}{.5\textwidth}
	\centering
	\includegraphics[scale=0.45]{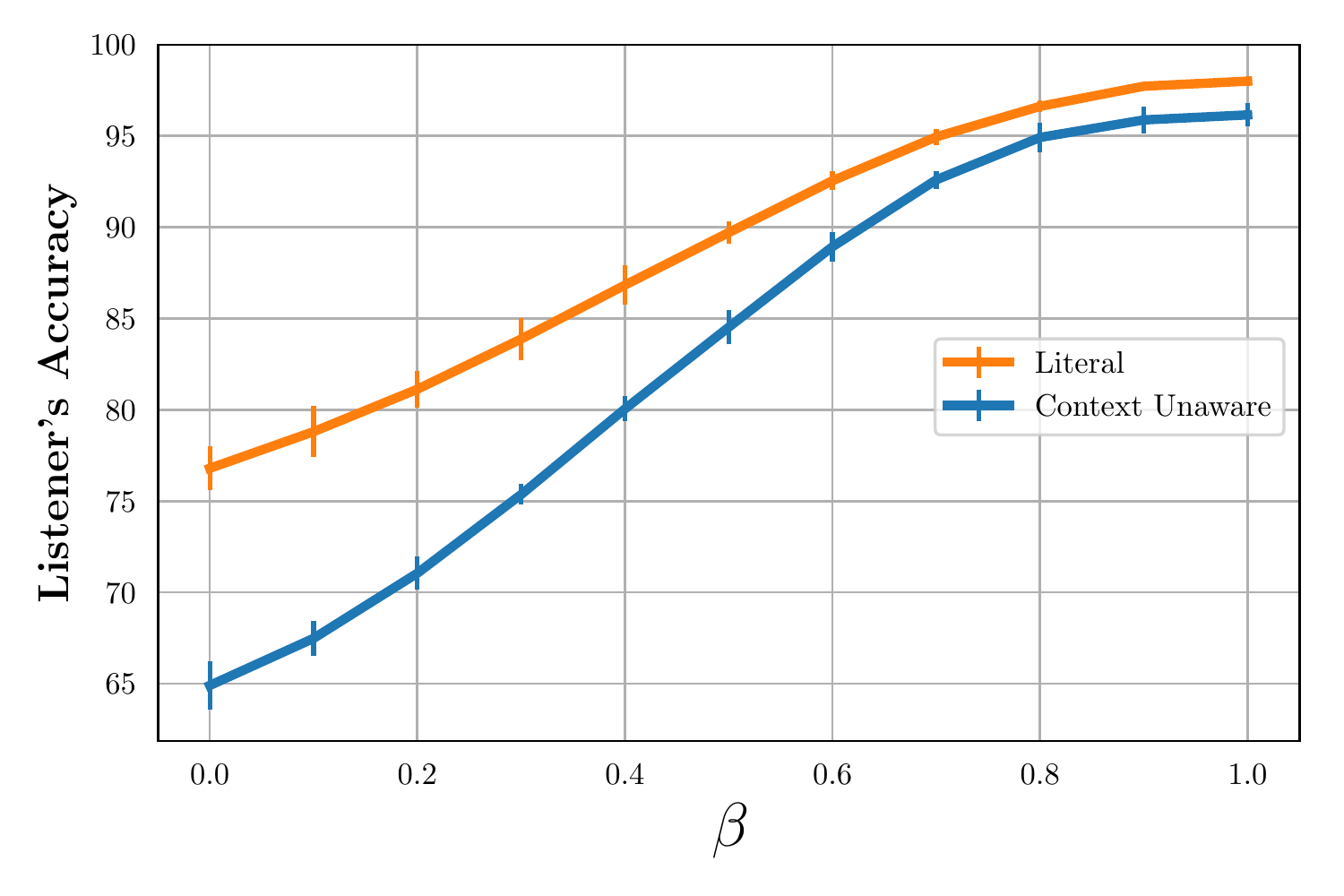}
	\caption{Effect of increasing listener's opinion ($\beta$).}
	\label{fig:beta_ablations}
	\end{subfigure}

\caption[width=\textwidth]{Left: Measuring the effect of using different length-penalty ($\alpha$) values to select the top-1 scoring utterance for context-unaware and pragmatic speakers for contexts of the \textit{object generalization} validation split (left). Right, measuring the the effect of various $\beta$-values used in turning the context-unaware and literal  speakers ($\beta=0.0$) to {\em pragmatic} speakers, under the optimal $\alpha$ of the left figure. In both plots, the y-axis reflects the performance of a listener who is used to rank {\em and} evaluate the utterances. Averages are with respect to 5 random seeds controlling the data splits and the initializations of the neural-networks. }	
\end{figure*}

To find the optimal length-penalty value ($\alpha$, Main Paper Eq.1) for image-based \textit{literal} and a \textit{context-unaware} speaker variants, we used our best-performing listener to simultaneously score and evaluate the utterances produced by the speakers for different values of $\alpha$ (Fig.~\ref{fig:alpha_ablations}). The best performing length penalty for a context-unaware speaker is $0.7$, and for a literal $0.6$. Given the optimal $\alpha$ values, for these models we show the effect of using different degrees of listener-awareness ($\beta$) in Fig.~\ref{fig:beta_ablations}. It is interesting to observe that even the context-unaware speaker can generate utterances that an evaluating listener can find them very discriminative, as long as is allows to rank them. 

\begin{figure*}
\centering
	\begin{subfigure}{.5\textwidth}
	\centering
	\includegraphics[scale=0.45]{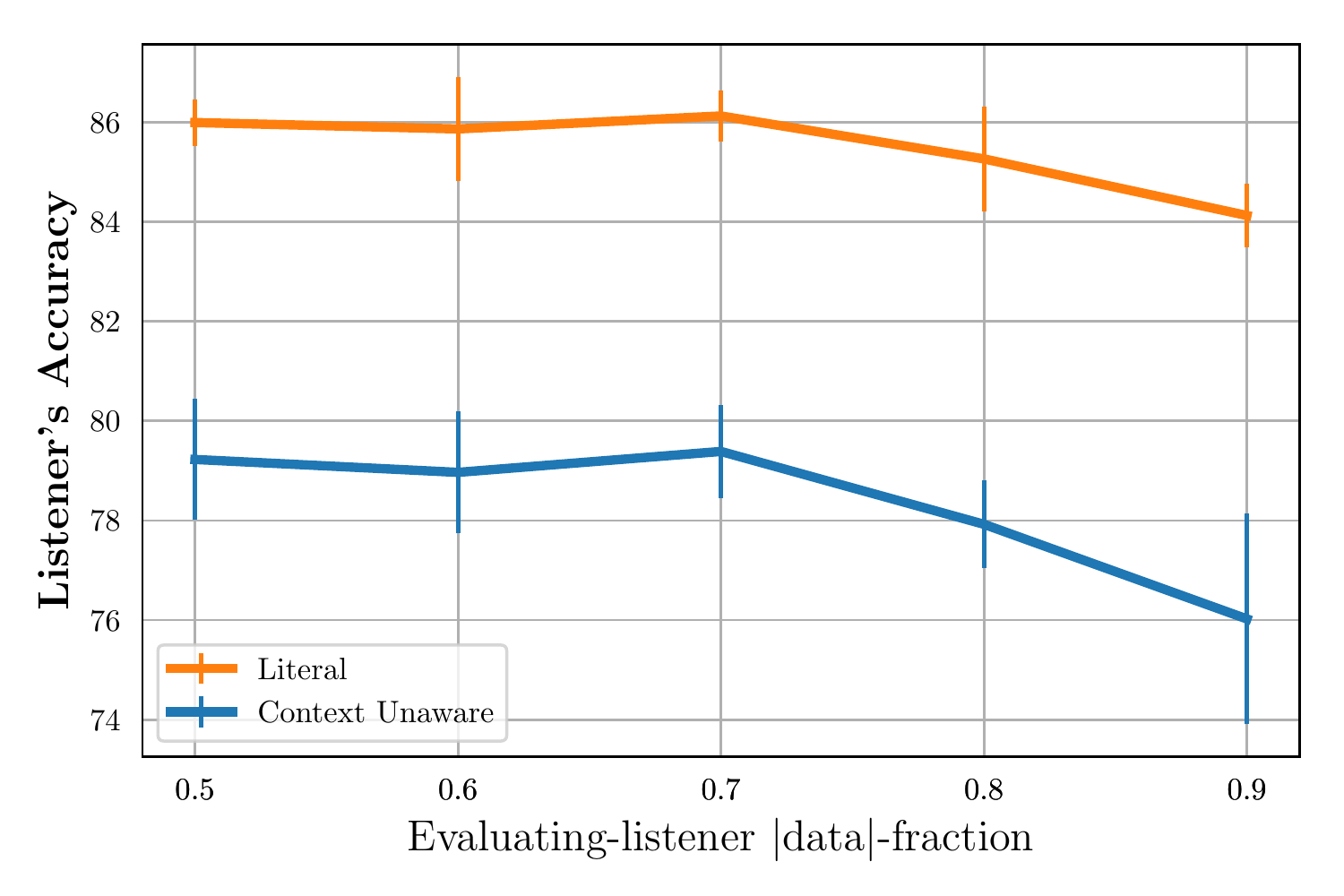}
	\caption{Speakers using a modest $\beta=0.5$ value.}
	\label{fig:listener_fraction_beta_5}
	\end{subfigure}%
	\begin{subfigure}{.5\textwidth}
	\centering
	\includegraphics[scale=0.45]{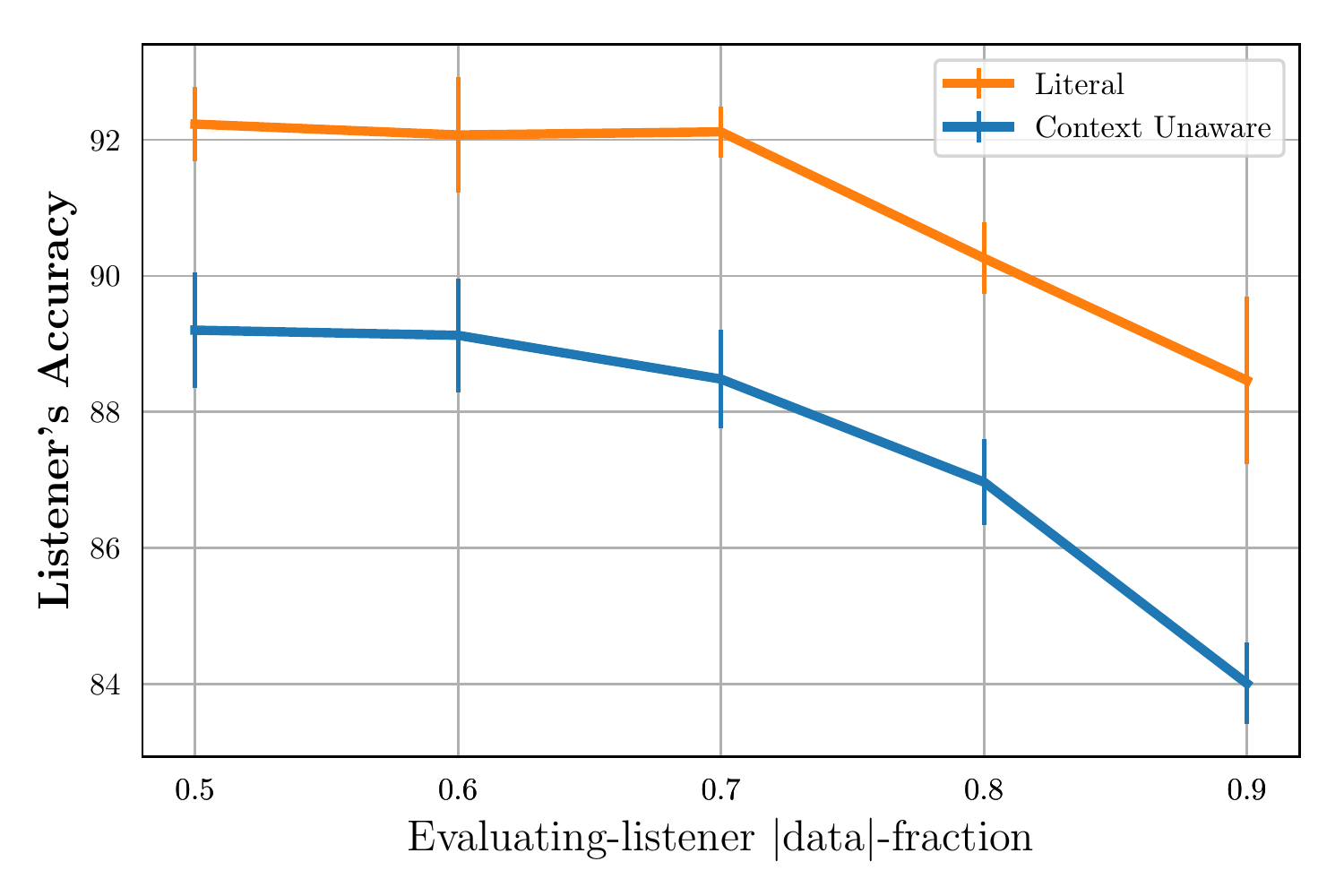}
	\caption{Speakers using the most aggressive $\beta=1.0$ value.}
	\label{fig:listener_fraction_beta_10}
	\end{subfigure}
% \caption{Effect of training data-size and neural-listener evaluation for {\em pragmatic} models with different degrees of listening awareness: $\beta=0.5$ (left), $\beta=1.0$ (right). The x-axis shows the fraction $f$ of the training data (80\% of the entire dataset) that was used the evaluating listener whose accuracy we report on the y-axis, for the top-1 generation of a pragmatic speaker. This speaker uses an 'internal' listener trained with the remaining $100-f\%$ of the training data. It appears that reducing the data for the 'internal' listener, while increasing them for the evaluating listener, slowly hurts overall performance. % Averages are with respect to 5 random seeds controlling the data splits and the initializations of the neural-networks, which in this case used image-based speakers and the best-performing listeners.
% }
\caption{Effect of partitioning the training data for the evaluating and `internal' listeners. Here, we turn context-unaware and literal speakers into pragmatic ones under two $\beta$ values. The x-axis shows the fraction ($f$) of the training data that was used to train the \textit{evaluating} listener (the remaining $100-f\%$ is used to train the \textit{internal} listener) of the resulting pragmatic speaker. On the y-axis we display the performance of the evaluating listener for the top-scoring model-generated utterance.}
% , under the test-split of the object generalization task.}
% Averages are with respect to 5 random seeds controlling the data splits and the initializations of the neural-networks, which in this case used image-based speakers and the best-performing listeners.
% (80\% of the entire dataset) is used for training.

% It appears that reducing the data for the 'internal' listener, while increasing them for the evaluating listener, slowly hurts overall performance. 

\label{fig:eval_listener_data_fraction}
\end{figure*}

In Fig.~\ref{fig:eval_listener_data_fraction} we demonstrate the effect that the relative (training) size of the evaluating listener vs. the 'internal' listener used by a {\em pragmatic} speaker has for the evaluating accuracy, for two values of $\beta$. In either case we observe a slow decline in evaluating accuracy as the training size for the evaluating listener increases (from 0.5 to 0.9) and consequently the training size for the 'internal' listener decreases (from 0.5 to 0.1). 

\subsubsection{Understanding out-of-class reference}

\label{ood_listening}
We complement the Table~\ref{table:ood_human_listening_without_attn} with the standard-deviations of the underlying accuracies in Table~\ref{table:ood_human_listening_without_attn_full}. We also report simple statistics regarding the underlying transfer classes in Table~\ref{table:ood_human_listening_basic_data_statistics}. We note that the transfer learning accuracies acquired by listeners operating with both point-clouds and images for these experiments were significantly lower ($\sim 7\%$ on average). We hypothesize that this is due to the fact that our (chair-trained) listener models that utilize point-clouds, rely on a pre-trained {\em single}-class PC-AE, unlike the pre-trained VGG (image encoder) which was fine-tuned with multiple ShapeNet classes. Also, for these experiments, [\texttildelow 1\% \texttildelow 7\%] (depending on the transfer class) of the tokens were not in the chair-vocabulary, and we chose to ignore them i.e.~treat them as white-space. Last, per Table~\ref{table:ood_human_listening_basic_data_statistics} in {\em all} transfer classes the \textit{with-part} population contains quite larger utterances than the \textit{without-part} (9.3 vs. 5.5 on average) and that even in the case of lamps, arguably the most dissimilar category from chairs, $20 + 37 = 57\%$ of the collected utterances are in the \textit{known} population.

% \input{../tables/appendix/out_of_dist_listening_with_attention.tex}
%!TEX root = ../../sections/main.tex
\begin{table*}[t!]
\centering
\begin{tabular}{|l||r|r|r|r|r|}
\hline
\backslashbox{Population}{\begin{tabular}{@{}c@{}}Class\end{tabular}} &  bed & chair & lamp  &  sofa & table\\
\hline
entire       & 56.4 $\pm 2.0\%$       &  77.4  $\pm 0.9\%$      &  50.1 $\pm 1.3\%$         &  53.6 $\pm 2.0\%$ &  63.7 $\pm 1.2\%$\\

\hline
known        & 55.8 $\pm 1.5\%$       & 77.8 $\pm 0.8\%$        &  51.9 $\pm 1.8\%$         &  55.0 $\pm 2.0\%$ &  65.5 $\pm 0.9\%$\\

\hline
with part    & {\bf 63.8} $\pm 4.2\%$ & 77.0 $\pm0.8\%$         & {\bf 60.3} $\pm  4.4\%$   & {\bf 55.1} $\pm 2.5\%$ &  {\bf 68.3} $\pm 2.6\%$\\

\hline
without part & 51.5 $\pm 3.0\%$      &  {\bf 80.5} $\pm 1.2\%$ & 47.1 $\pm 2.8\%$           &  54.7 $\pm 5.5\%$ &  62.7 $\pm 0.9\%$\\
\hline
\end{tabular}

\caption{Transfer-learning of neural listeners in novel object {\em classes}: average accuracies {\em with} standard deviations (complementing Table~\ref{table:ood_human_listening_without_attn_full}, Main Paper). The sub-populations denote \textit{entire}: all collected utterances, \textit{known}: utterances containing {\em only} chair-training-vocabulary words, \textit{with-part}: subset of \textit{known}, with utterances containing at least one part-related word, \textit{without-part} subset of \textit{known} and complement of \textit{with-part}. For reference the test-chair statistics are shown (first row) but not included in the reported average (last row). }
\label{table:ood_human_listening_without_attn_full}
\end{table*}
% \clearpage
% \subsection{Further qualitative results}

\begin{figure*}
\centering
\caption{\textbf{Examples of attention weights on human utterances}. The listener's LSTM appears to learn attention weights that emphasize the more informative words disambiguating the referent. For these results the  \listenerA\ listener is used and the attention-scores are extracted when the target object is grounding the LSTM.}
\includegraphics[width=\textwidth]{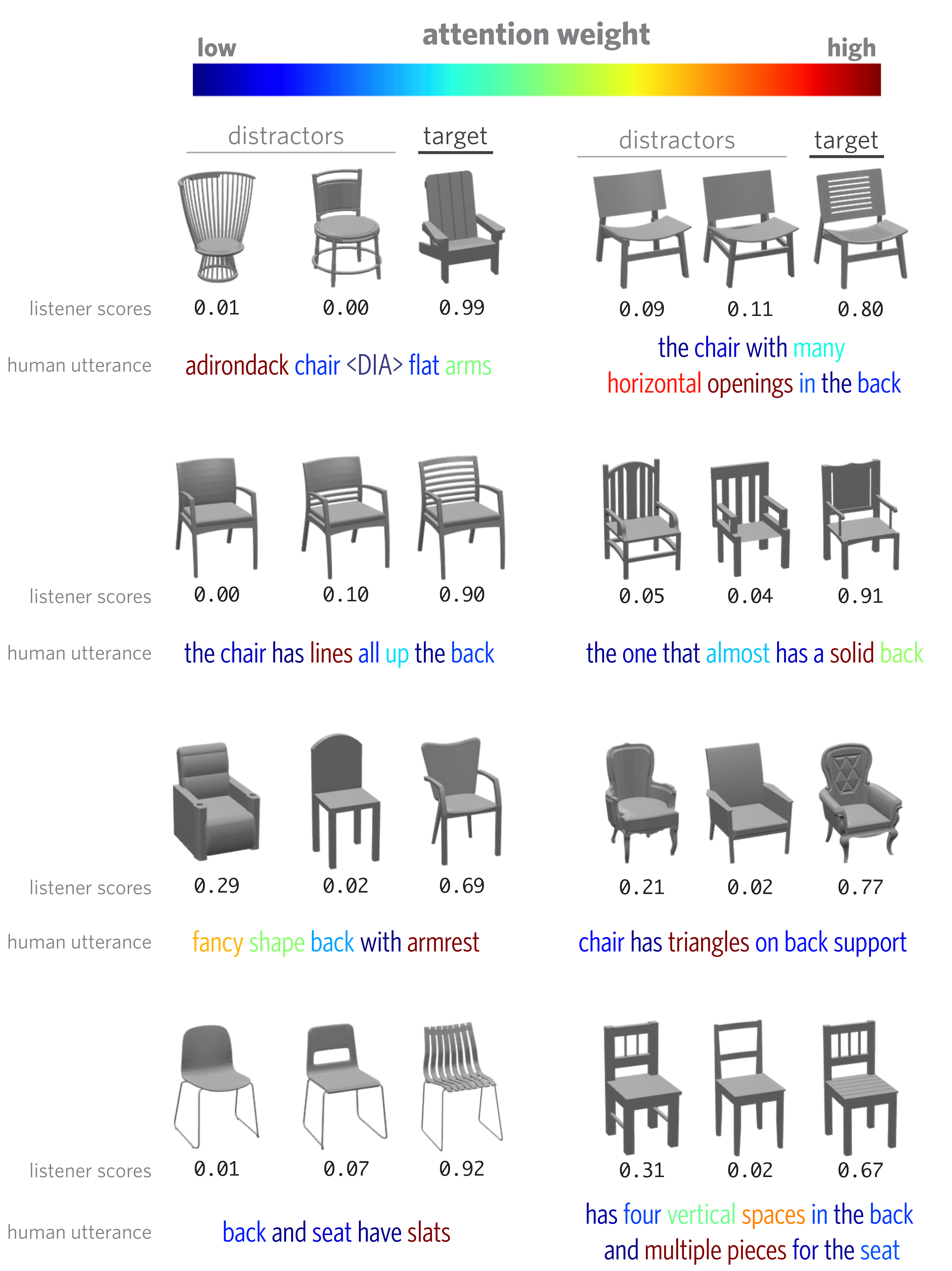}
\label{fig:human_utterances_and_attention}
\end{figure*}

\begin{figure*}[tb]
\centering
\caption{\textbf{Examples of lesioning all but the mentioned part}. Here, we show the response of a \listenerA\  listener tested with visual representations of entire objects (left column, three chairs) vs. its response when it receives
{\bf only} the visual features corresponding to the referred semantic-part (right column). The corresponding utterance is shown left-most of each row. In these examples the listener assigns higher confidence to the actual target when the isolated parts are considered instead of the entire objects, implying that further performance gains can occur with an explicit part-aware visual attention mechanism.}
\includegraphics[width=\textwidth]{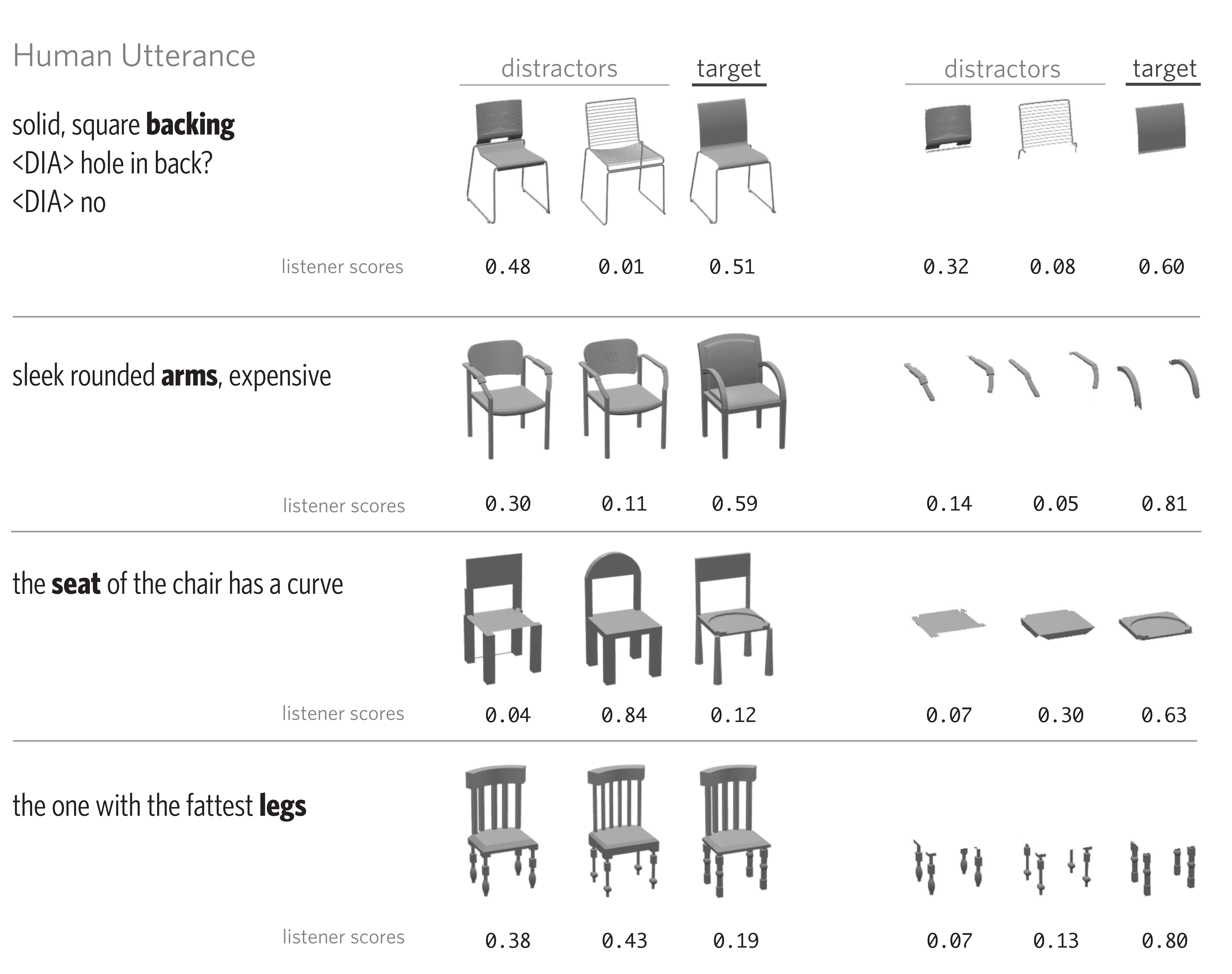}
\label{fig:listening_with_single_part_only}
\end{figure*}

\begin{figure*}[tb]
\centering
\caption{\textbf{Pragmatic vs. literal speakers for two modalities}. More examples of pragmatic vs. literal generations in Hard contexts. Tor-row includes examples from image-based speakers. Bottom-row from point-based ones.}
\includegraphics[width=\textwidth]{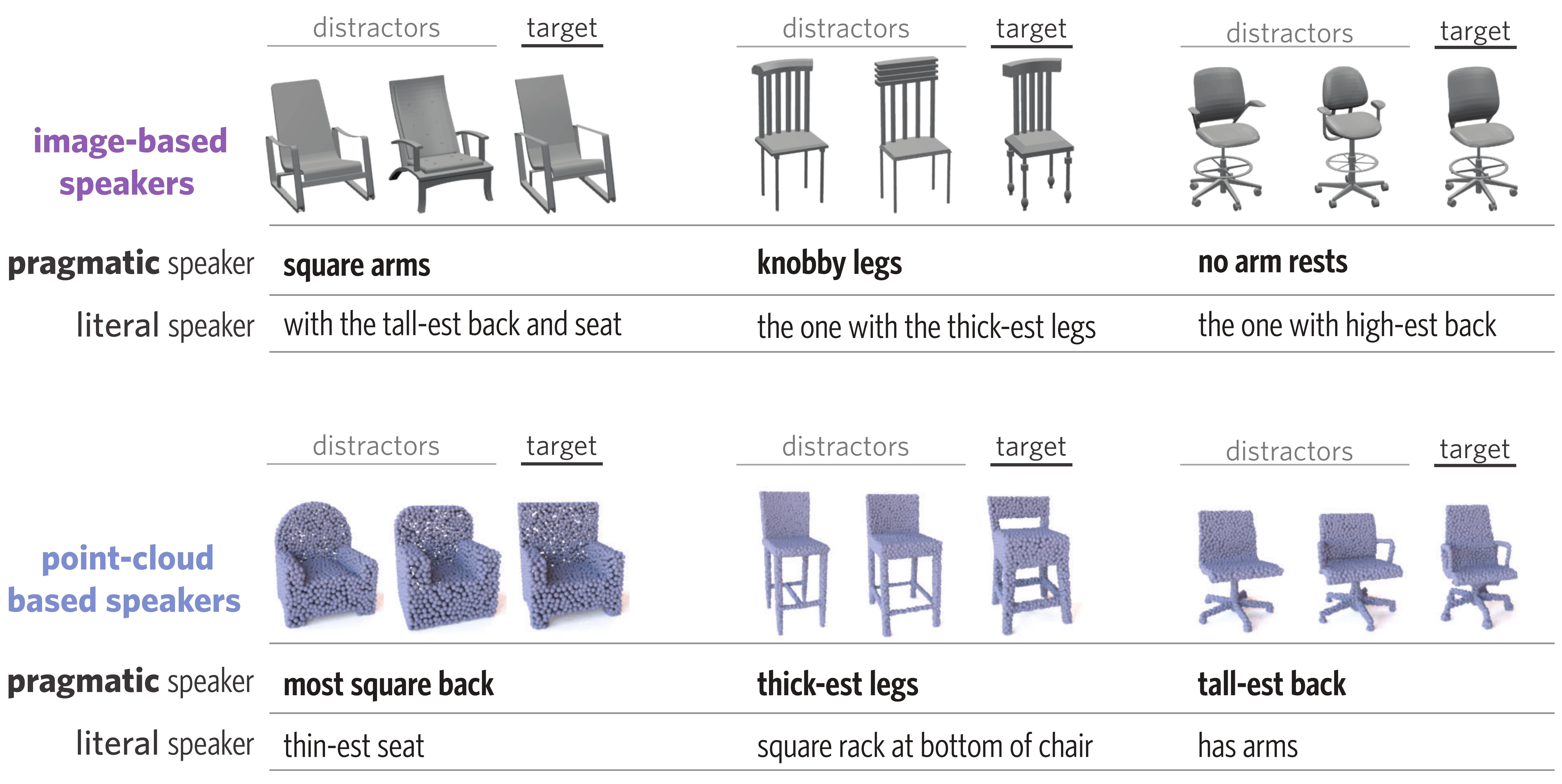}
\label{fig:pragma_vs_literal_hard}
\end{figure*}

\begin{figure}
  \begin{subfigure}{\textwidth}  	
    \caption{\textbf{Model generations with real images.} The top-scoring utterance of a pragmatic model is displayed under  each context.}   
    \centering\includegraphics[scale=0.5, width=\textwidth]{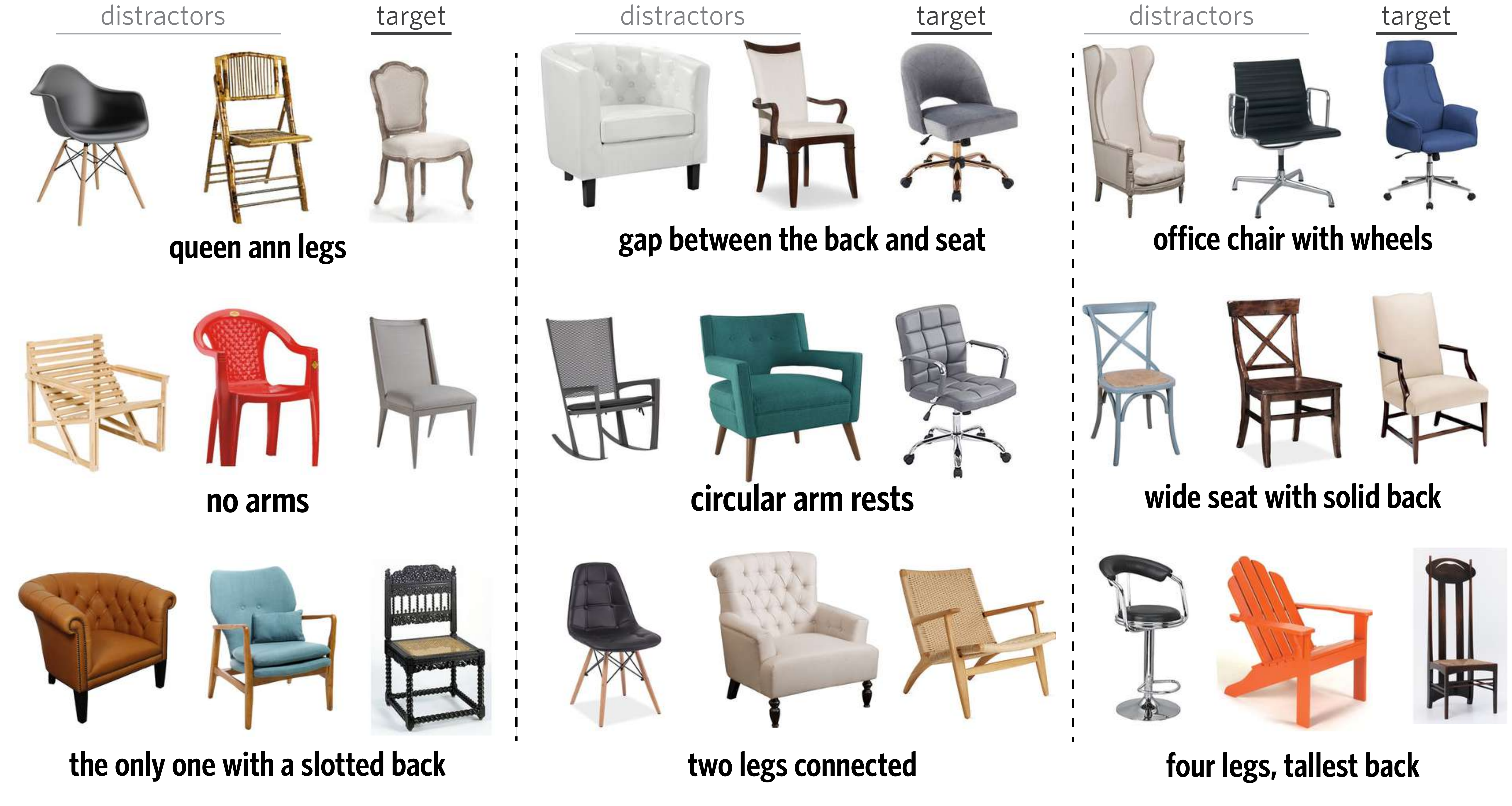}    
  \end{subfigure}
  \begin{subfigure}{\textwidth}
  	\vspace{30pt}

    \caption{\textbf{Human-utterance comprehension with unseen object classes.} The human utterance is color-coded according to the attention placed by a chair-trained listener who also evaluates the object-utterance compatibility (scores shown under its context).}
    \centering\includegraphics[scale=0.5, width=\textwidth]{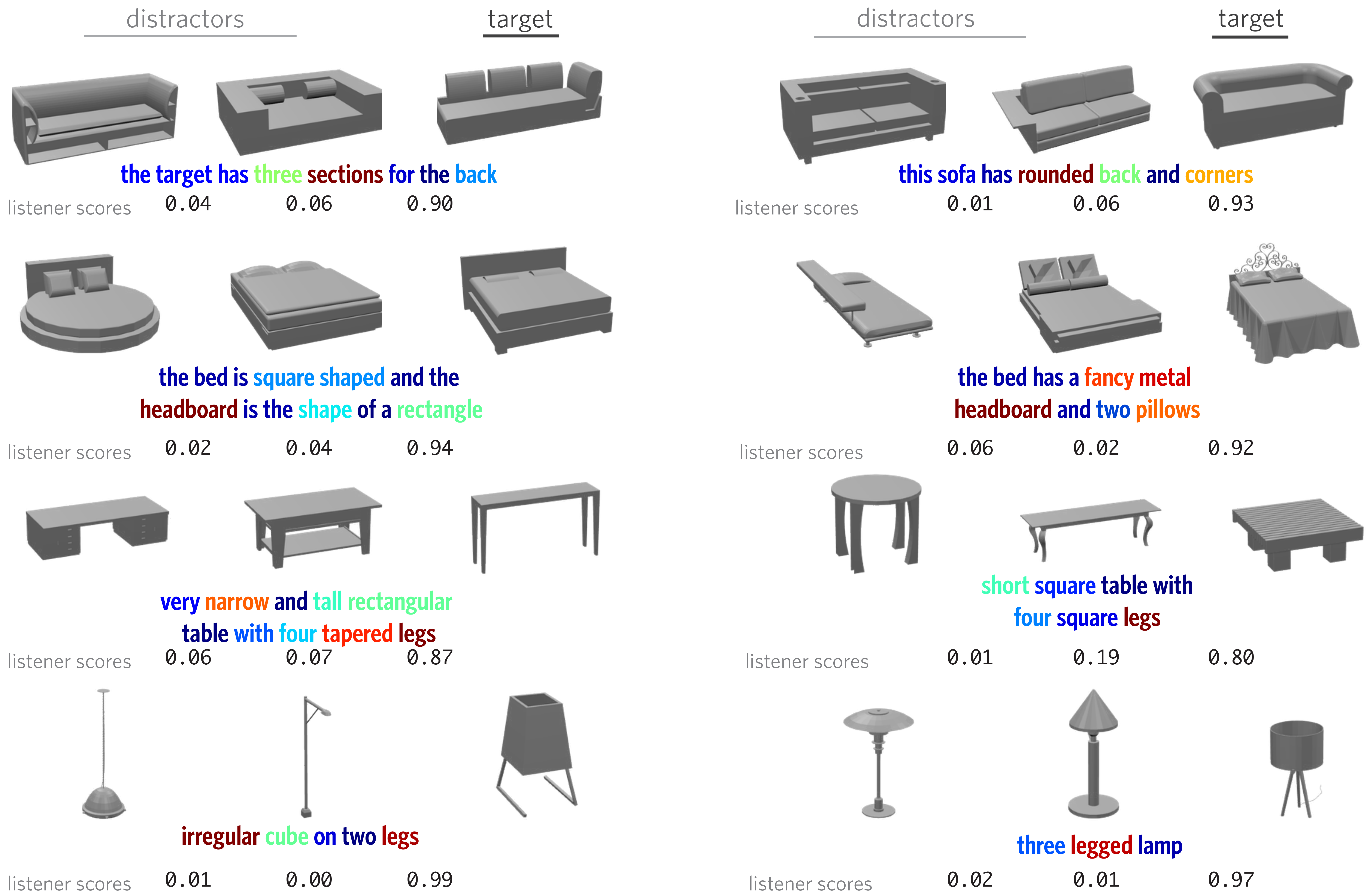}    
  \end{subfigure} 
\end{figure}
\newpage
% \begin{figure*}
% \centering
% \caption{\textbf{Speaking in \textit{novel} classes.} In orange-color are \textit{model}-generations of a chair-speaker describing the \textit{non-chair} object above it (inside each orange box). Here, we first use a chair-listener to create easily-separable communication contexts to which we then apply our speaker. Concretely, the listener scores the utterance-object compatibility of \textit{all} ShapeNet objects of a given class e.g.~table, under a `query' (utterance) e.g.~`modern'. We use the top-5 scoring and least-5 scoring objects (shown in left/right panels of each row respectively) to select at random a {\it target} (orange box) from the former set and two {\it distractors} (cyan boxes) from the latter. In the resulting communication context, we apply our speaker and display in orange-color the generation. The queries used for this experiment are shown in the left-most part of the Figure.}
% \includegraphics[width=\textwidth]{../figures/appendix/rank_ans_speak_out_of_class.pdf}
% \label{fig:human_utterances_and_attention}
% \end{figure*}

\begin{landscape}
\begin{figure}[h]
\centering
\caption{\textbf{Effect of context on production}: Synthetic utterances generated by a \textit{literal} and \textit{pragmatic} image-based speaker. The top and bottom rows show utterances produced for the same target in a Easy and Hard context, respectively. The \listenerA\ (with point-clouds and images and attention) listener is used to predict the target and its confidence is displayed above each utterance. While both speaker models produce similarly effective utterances in Easy contexts, the literal speaker fails to produce effective utterances in Hard contexts.}
\includegraphics[scale=0.25]{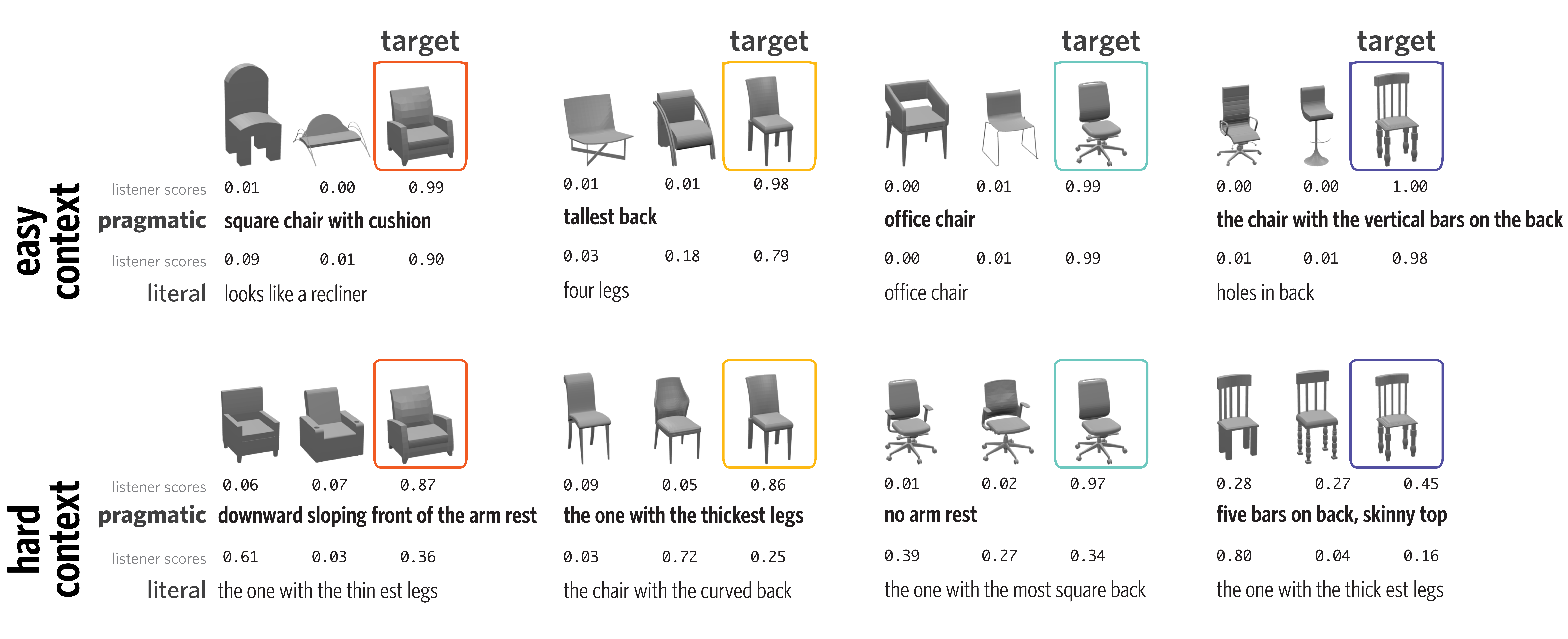}
\label{fig:context_effects}
\end{figure}
\end{landscape}

%%% Failure Listening and Speaking Cases
\begin{figure}
  \begin{subfigure}{\textwidth}
    \caption{\textbf{Neural-listener failure cases}. Our top-performing listener model appears to struggle to interpret referential language that relies on metaphors, precisely counting parts, or (to a less degree) negations. All examples are drawn from the test set and were correctly classified by human listeners in the original task.}
    \vspace{-10pt}
	\label{fig:listener_errors}	
    \centering\includegraphics[scale=0.5, width=\textwidth]{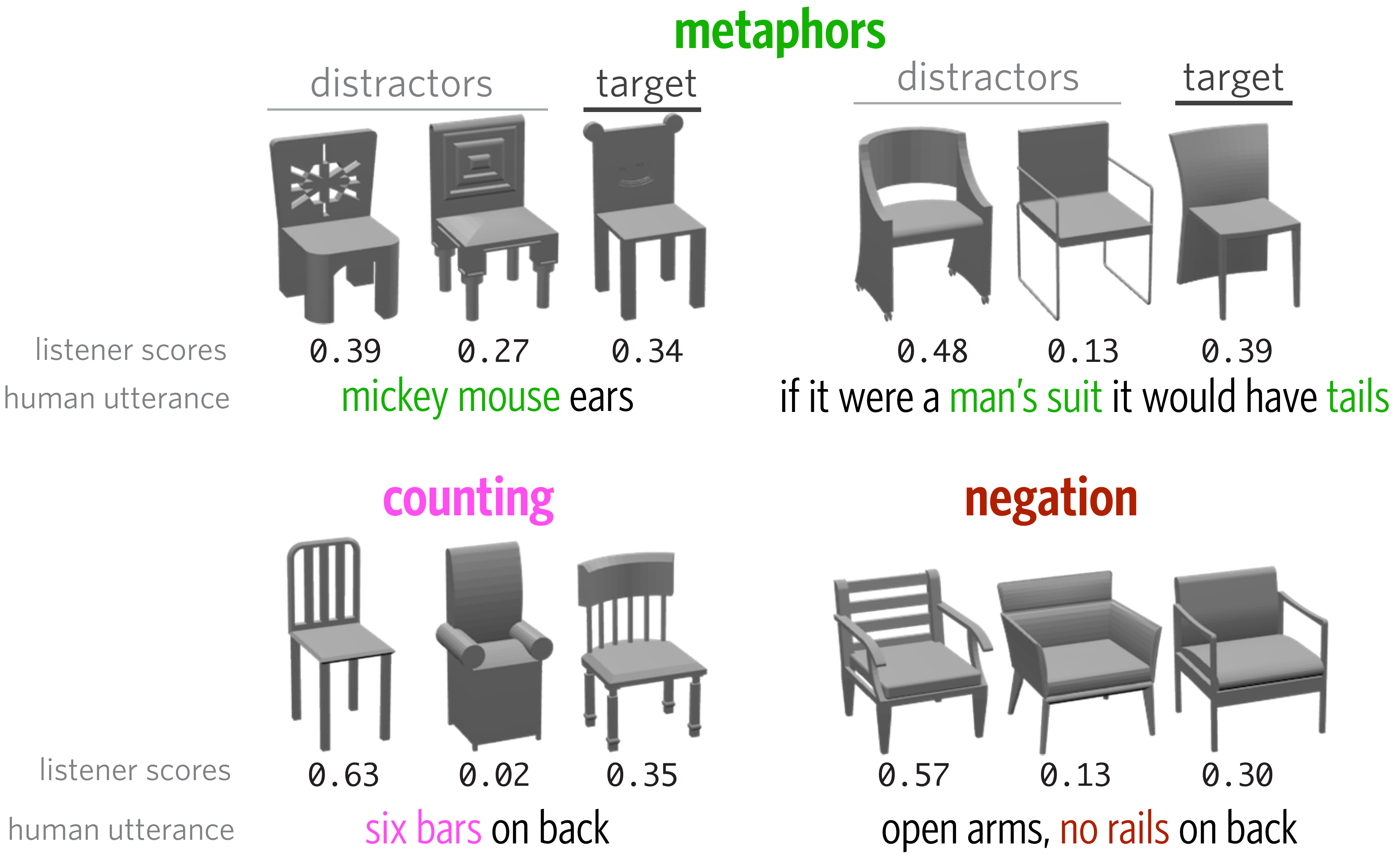}    
  \end{subfigure}
  \begin{subfigure}{\textwidth}
  	\vspace{30pt}
    \caption{\textbf{Neural-speaker failure cases}. Sometimes even the \textit{pragmatic} speaker produces insufficiently specific utterances that mention only undiagnostic features, or produces utterances that are literally false of the target (e.g.~there technically \emph{is} a hole in the back) while still succeeding in distinguishing the objects.}
	\label{fig:speaker_errors}
    \centering\includegraphics[scale=0.5, width=\textwidth]{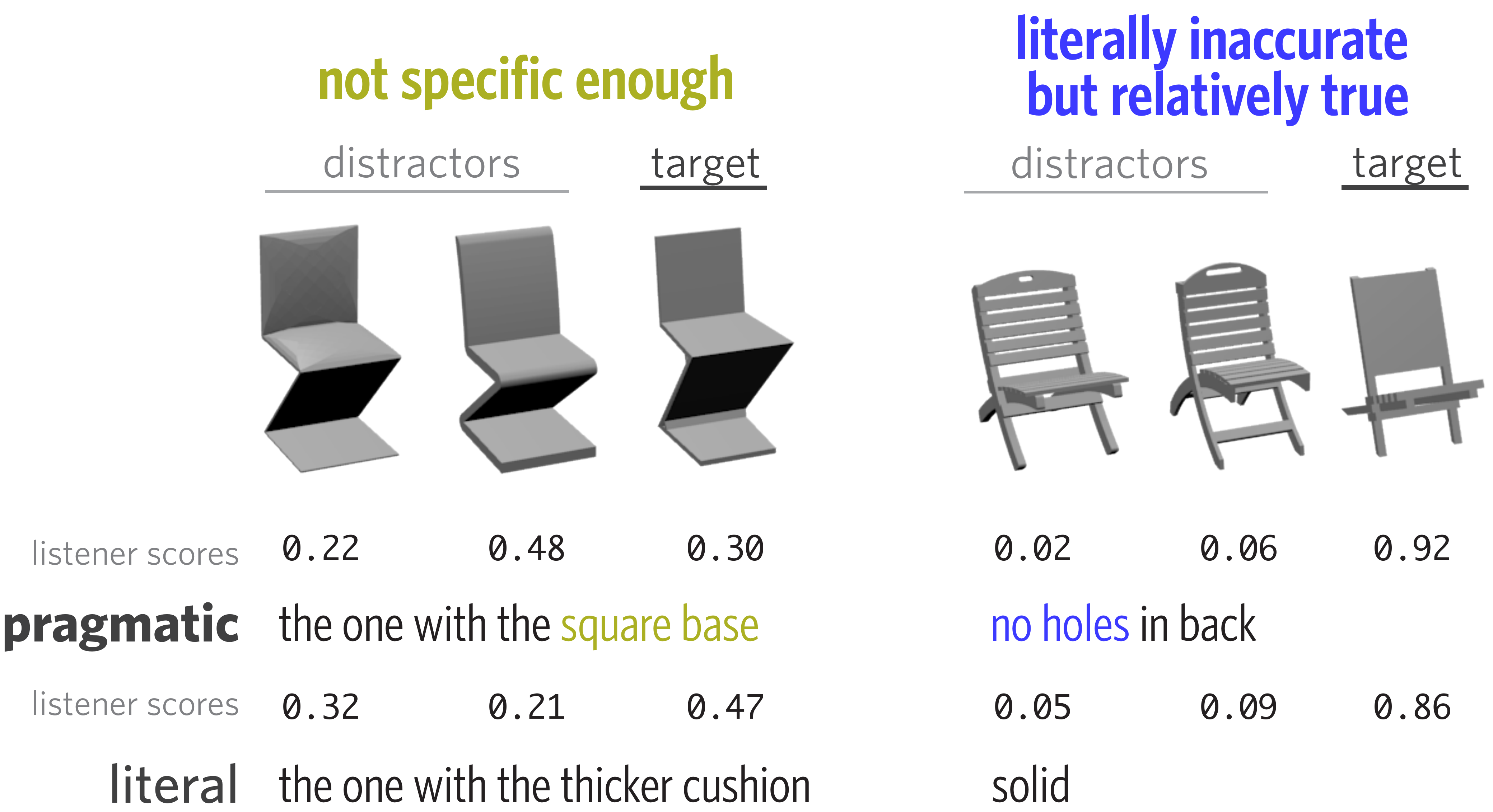}    
  \end{subfigure} 
\end{figure}

\clearpage 
\newpage
\subsection{Miscellaneous}

%!TEX root = ../../sections/main.tex
% \begin{table*}[tbh!]
\begin{table}[tbh!]
\centering
\resizebox{\textwidth}{!}{
\begin{tabular}{|l|l|l|l|l|l|l|l|l|l|l|l|}
\hline
\multirow{2}{*}{\textbf{Easy}}   & \textbf{word} & office & sofa      & regular   & folding & wooden & stool    & wheels  & metal   & normal  & rocking \\ \cline{2-12}
                                & \textbf{pmi}  & -1.70  & -0.94     & -0.88     & -0.84   & -0.83  & -0.79    & -0.78   & -0.71   & -0.67   & -0.66   \\ \hline
\multirow{2}{*}{\textbf{Hard}} & \textbf{word} & alike  & identical & thickness & texture & darker & skinnier & thicker & perfect & similar & larger  \\ \cline{2-12}
                                & \textbf{pmi}  & 0.69   & 0.67      & 0.67      & 0.66    & 0.65   & 0.64     & 0.63    & 0.62    & 0.62    & 0.61    \\ \hline
\end{tabular}
}

\centering
% \captionsetup{indention=20cm, width=\textwidth}
\caption{Most distinctive words in each context type according to point-wise mutual information (excluding tokens that appeared fewer than 30 times in the dataset). Lower numbers are more distinctive of Easy and higher numbers are more distinctive of Hard.}
\label{table:pmi}
\end{table}

\begin{figure}[h]
	\centering	
	\includegraphics[width=.5\textwidth]{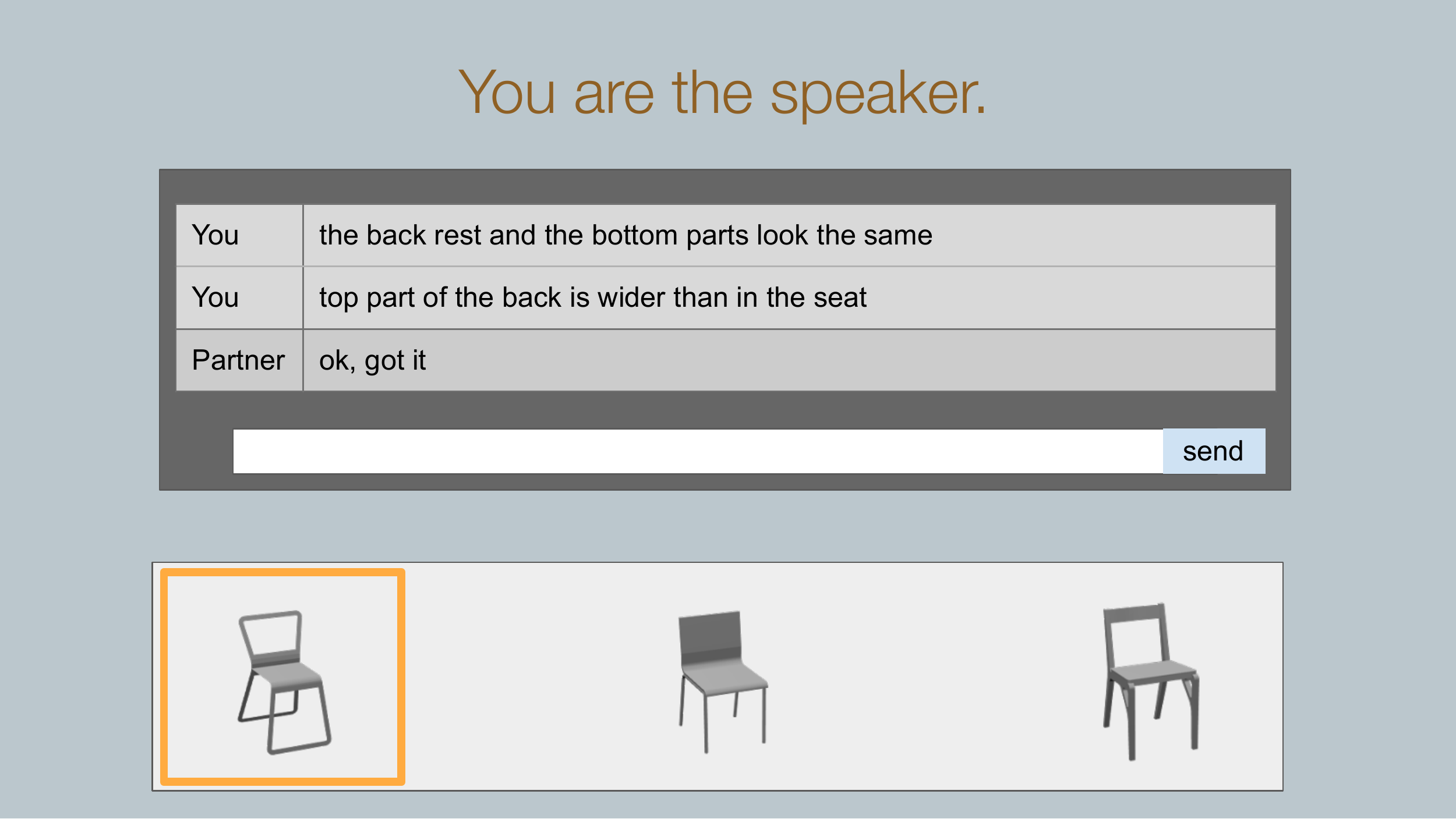}
	\caption{Reference game interface. Communication was natural without any system constraints being imposed.}
	\label{fig:game_ui}
\end{figure}

Each game consisted of 69 trials (unique triplets) and participants swapped speaker and listener roles with the conclusion of each trial. The game's interface is depicted in Figure~\ref{fig:game_ui}. Participants were allowed to play multiple games, but most participants in our dataset played exactly one game (81\% of participants). The most distinctive words in each triplet type (as measured by point-wise mutual information) are shown in Table \ref{table:pmi}).

\begin{figure}[h]
\centering
\includegraphics[width=.5\textwidth]{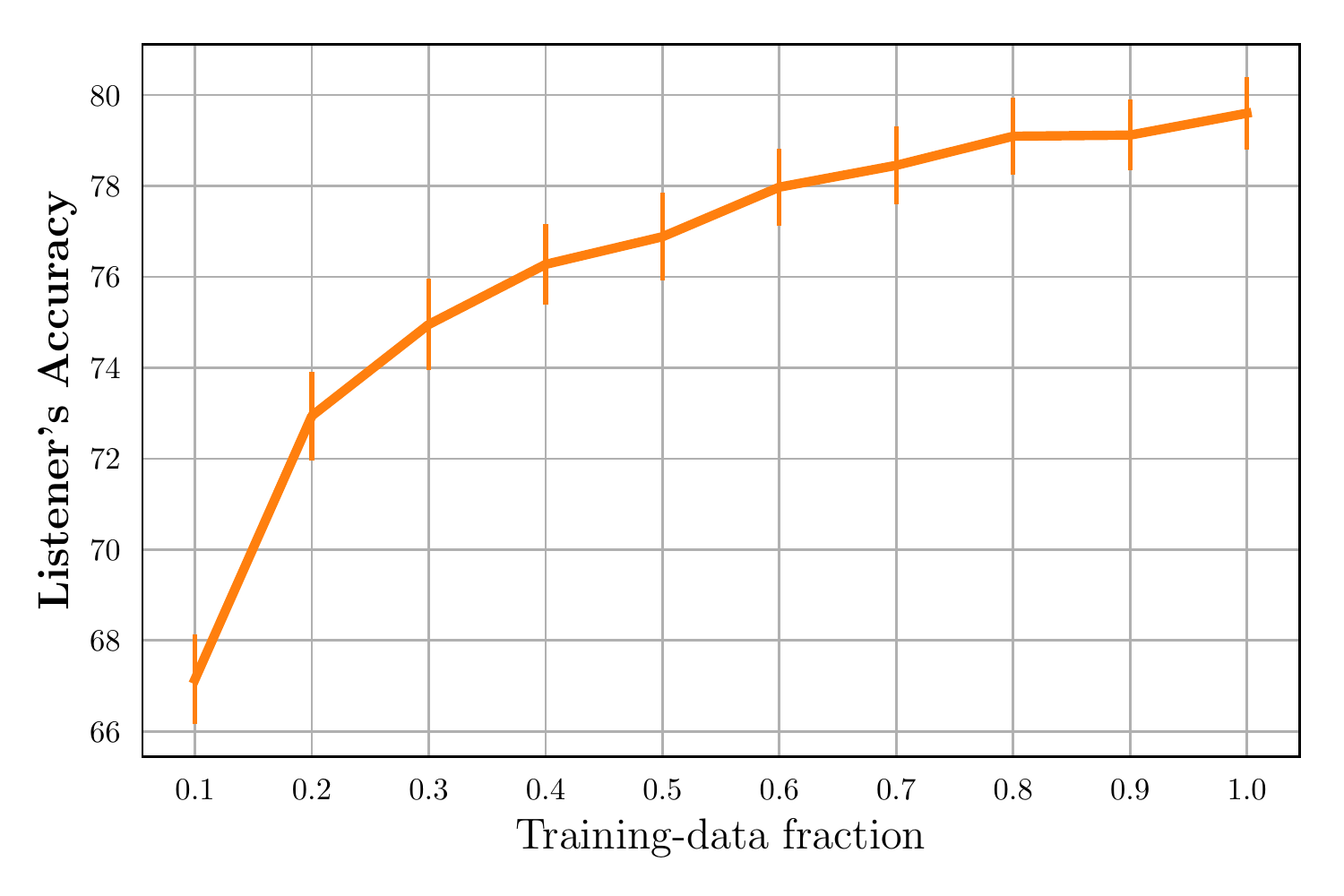}
\caption{Listener's accuracy for different sizes of training data, under the {\em object} generalization task. The original split includes [80\%, 10\%, 10\%] for training/test/val purposes, thus the maximum size of training data is $0.8$ of the entire dataset corresponding to the value (fraction) 1.0 in the x-axis.  The listener model uses the \listenerA\ architecture with word attention, images and point-clouds and its accuracy is measured on the original ($10\%$) test split. Results are averages of 5 random seeds controlling the original data split and the neural-net's initialization.}
\label{fig:training_sizes}
\end{figure}

\end{document}